\documentclass[10pt,journal,compsoc]{IEEEtran}

\ifCLASSOPTIONcompsoc
  \usepackage[nocompress]{cite}
\else
  \usepackage{cite}
\fi
\usepackage{graphicx}
\usepackage{amsmath}
\usepackage{amssymb}
\usepackage{booktabs}
\usepackage{bbding}
\usepackage{epsfig}
\usepackage{multirow}
\usepackage{threeparttable}  
\usepackage{color}
\usepackage{array}
\usepackage{makecell}
\usepackage{stfloats}
\usepackage[table]{xcolor}
\usepackage[misc]{ifsym}
\usepackage[accsupp]{axessibility}
\usepackage{hyperref}

\ifCLASSINFOpdf
\else
\fi

\begin{document}

\title{MixFormer: End-to-End Tracking with \\ Iterative Mixed Attention}

\author{Yutao~Cui, Cheng~Jiang,
Gangshan~Wu,~\IEEEmembership{Member,~IEEE} and~Limin~Wang,~\IEEEmembership{Member,~IEEE}% <-this % stops a space
\IEEEcompsocitemizethanks{\IEEEcompsocthanksitem Yutao Cui, Cheng Jiang, Gangshan Wu and Limin Wang are with the State Key Laboratory for Novel Software Technology, Nanjing University, Nanjing, China, 210023 (E-mail:cuiyutao@smail.nju.edu.cn; mg1933027@smail.nju.edu.cn; gswu@nju.edu.cn; lmwang@nju.edu.cn).
}% <-this % stops a space
}

% \markboth{Journal of \LaTeX\ Class Files,~Vol.~14, No.~8, August~2015}%
% {Shell \MakeLowercase{\textit{et al.}}: Bare Advanced Demo of IEEEtran.cls for IEEE Computer Society Journals}

\IEEEtitleabstractindextext{%
\begin{abstract}
Visual object tracking often employs a multi-stage pipeline of feature extraction, target information integration, and bounding box estimation. To simplify this pipeline and unify the process of feature extraction and target information integration, in this paper, we present a compact tracking framework, termed as {\em MixFormer}, built upon transformers. Our core design is to utilize the flexibility of attention operations, and propose a Mixed Attention Module (MAM) for simultaneous feature extraction and target information integration. This synchronous modeling scheme allows to extract target-specific discriminative features and perform extensive communication between target and search area. Based on MAM, we build our MixFormer trackers simply by stacking multiple MAMs and placing a localization head on top.
Specifically, we instantiate two types of MixFormer trackers, a hierarchical tracker {\em MixCvT}, and a non-hierarchical tracker {\em MixViT}.
For these two trackers, we investigate a series of pre-training methods and uncover the different behaviors between supervised pre-training and self-supervised pre-training in our MixFormer trackers. We also extend the masked pre-training to our MixFormer trackers and design the competitive TrackMAE pre-training technique.
Finally, to handle multiple target templates during online tracking, we devise an asymmetric attention scheme in MAM to reduce computational cost, and propose an effective score prediction module to select high-quality templates. 
Our MixFormer trackers set a new state-of-the-art performance on seven tracking benchmarks, including LaSOT, TrackingNet, VOT2020, GOT-10k, OTB100 and UAV123. 
In particular, our MixViT-L achieves AUC score of 73.3\% on LaSOT, 86.1\% on TrackingNet, EAO of 0.584 on VOT2020, and AO of 75.7\% on GOT-10k.
Code and trained models are publicly available at 
\href{https://github.com/MCG-NJU/MixFormer}{https://github.com/MCG-NJU/MixFormer}.
\end{abstract}

}

\maketitle

\IEEEdisplaynontitleabstractindextext

\IEEEpeerreviewmaketitle

\ifCLASSOPTIONcompsoc
\IEEEraisesectionheading{\section{Introduction}\label{sec:introduction}}
\else
\section{Introduction}
\label{sec:introduction}
\fi

\IEEEPARstart{V}{isual} object tracking~\cite{updt,mdnet,dcf_,CRPN,bacf,updt,staple,vital} has been a long-standing problem in computer vision area for decades, which aims to estimate the state of an arbitrary target in video sequences given its initial state. It has been deployed in various applications such as human computer interaction~\cite{introduction1} and visual surveillance~\cite{introduction2}. However, so far how to design a simple yet effective end-to-end tracker is still challenging in the real-world scenarios. The main challenges are from aspects of scale variations, object deformations, occlusion, and confusion from similar objects.

\begin{figure}[t]
\centering
\includegraphics[width=\linewidth]{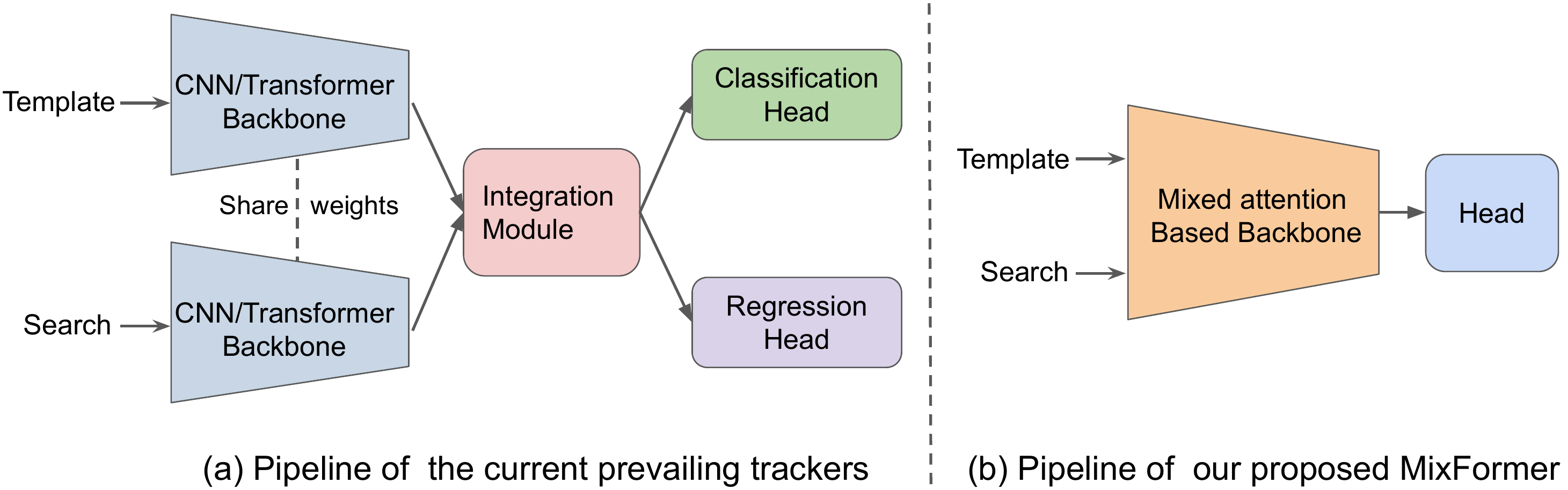}
\caption{{\bf Comparison of tracking pipeline}. (a) The dominant tracking framework contains three components: a convolutional or transformer backbone, a carefully-designed integration module, and task-specific heads. (b) Our MixFormer is more compact and only composed of two components: a target-search mixed attention based backbone and a simple localization head. 
}
\label{fig:motivation}
\end{figure}

Current prevailing trackers~\cite{siamfc,siamrpn,siamban,ocean,dcf_,atom,dimp,fcot,tt,tmt,stark} typically employ a multi-stage pipeline as shown in Fig.~\ref{fig:motivation}. It contains several components to accomplish the tracking task: (1) a {\bf backbone} to extract generic features of tracking target and search area, (2) an {\bf integration module} to allow information communication between tracking target and search area for subsequent target-aware localization, (3) task-specific {\bf heads} to precisely localize the target and estimate its bounding box. Integration module is the key of tracking algorithms as it is responsible for incorporating the target information to bridge the steps of generic feature extraction and target-aware localization. 
Traditional integration methods include correlation-based operations (e.g. SiamFC~\cite{siamfc}, SiamRPN~\cite{siamrpn}, CRPN~\cite{CRPN}, SiamFC++~\cite{siamfc++}, SiamBAN~\cite{siamban}, OCEAN~\cite{ocean}) and online learning algorithms (e.g., DCF~\cite{dcf_}, KCF~\cite{kcf}, CSR-DCF~\cite{csr_dcf}, ATOM~\cite{atom}, DiMP~\cite{dimp}, FCOT~\cite{fcot}). Recently, thanks to its global and dynamic modeling capacity, Transformers~\cite{transformer} are introduced to perform attention based integration and yields good tracking performance (e.g., TransT~\cite{tt}, TMT~\cite{tmt}, STMTrack~\cite{stmtrack}, TREG~\cite{treg}, STARK~\cite{stark}, DTT~\cite{dtt}). 
However, these transformer based trackers still depend on the CNN for generic feature extraction, and simply apply attention operations for target information integration in the latter high-level and abstract representation space. We analyze that these CNN representations are limited as they are typically pre-trained for generic object recognition and might neglect finer structure information for tracking. In addition, these CNN representations employ local convolutional kernels and are ineffective in global modeling. Therefore, the decoupled CNN representation and integration modules still fail to fully unleashing power of attention operations (Transformers) in the whole tracking pipeline.

To overcome the above issues, we present a new perspective on tracker design by unifying visual feature extraction and target information integration within a single module. This unified processing paradigm shares several key advantages. First, it enables the feature extraction to be aware of the corresponding tracking target and thus capture more target-specific features. Second, it also allows the target information to be more extensively integrated into search area, and thereby better to capture their correlation. In addition, as shown in Fig.~\ref{fig:motivation}, this will result in a more compact and efficient tracking pipeline only with a single backbone and a tracking head, without an explicit integration module.

To this end, in this paper, we present the \emph{MixFormer}, a neat tracking framework designed for unifying the feature extraction and target integration  with a transformer-based architecture. Attention module is a very flexible architectural building block with dynamic and global modeling capacity, which makes few assumption about the data structure and could be applied for general relation modeling. Our core idea is to utilize this flexibility of attention operation, and devise the {\em mixed attention module} (MAM) that performs both feature extraction and information interaction of target template and search area at the same time. Specifically, in our MAM, we propose a hybrid interaction scheme with both self-attention and cross-attention operations on the tokens from target template and search area. The self-attention is responsible to extract their own features of target template or search area, while the cross-attention allows for the communications between them to mix their information. 
To further reduce the computational cost of MAM and allow for multiple templates to handle object deformation, we further present a customized {\em asymmetric} attention scheme by pruning the unnecessary target-to-search area cross-attention. 
This enables our MixFormer trackers to be easily adapted for multiple target template inputs. In practice, we propose a score based target template update mechanism, to select reliable and high-quality online templates.

We instantiate two types of MixFormer trackers, a hierarchical tracker with progressive downsampling and depth-wise convolutional projections (termed as {\bf MixCvT}), and a non-hierarchical tracker built upon the plain backbone of ViT (termed as {\bf MixViT}).
For the former one, motivated by the success of hybrid convolutional and transformer backbones~\cite{cvt, pvt, localvit, t2t, swin}, we want to introduce the inductive bias in the tracker design by using a progressive downsampling architecture and introducing locality modeling in the Wide Mixed Attention Module (W-MAM). 
Specifically, we use the CvT~\cite{cvt} backbone to  build our hierarchical MixFormer tracker MixCvT by stacking the layers of Convolutional Patch Embedding and W-MAM and finally placing a simple localization head on top. 
Due to the progressive downsampling and locality modeling in MixCvT, it exhibits desirable properties (i.e., shift, scale and distortion invariance) of CNNs~\cite{cvt} and achieves promising performance. However, these adaptations yield a hierarchical architecture with a tremendous amount of tokens in the first two stages and relatively complex computational steps. 

In the other instantiation, we explore a~{\em more general},~\emph{simpler} and~\emph{non-hierarchical} backbone of Vision Transformer (ViT)~\cite{vit} for designing MixFormer tracker. Motivated by its design simplicity and promising results of ViT in image and video recognition, we aim to build a plain MixFormer tracker MixViT without inductive bias for high modeling flexibility and tracking efficiency. We construct the MixViT tracker with a simple patchification scheme and the Slimming Mixed Attention Module (S-MAM) by removing the convolutional projection in Wide MAM. Specifically, we first slice the image into non-overlapping patches and project them into a token sequence with patch embeddings and position embeddings. Then we stack multiple S-MAMs on the fixed-length token sequence and finally place a pyramidal localization head.
For these two types of MixFormer trackers, we design customized localization heads for accurate bounding box estimation. Especially, we design a pyramidal corner head for non-hierarchical MixViT, which can supplement some multi-scale information.

Due to the small size of tracking datasets and the overall complexity of tracking task, pre-training is an necessary step for training effective transformer-based trackers. The existing successful trackers~\cite{siamfc,siamrpn,siamban,ocean,treg,transt,stark} often uses the ImageNet pre-trained models to initialize the weights of their feature extraction module. Although we mix the tokens from the target template and search area in our MAM, we could directly apply the ImageNet pre-trained CvT or ViT models as the initialization of the corresponding trackers, thanks to the flexibility of attention operations. In practice, for MixCvT, we empirically study the supervised pre-training on ImageNet-1k and ImageNet-22k, demonstrating that pre-trained model with more semantic supervision also contributes to a better tracking performance. For MixViT, due to its less inductive bias and higher flexibility, the ViT unlocks the capability of dropping image patch tokens and unleash the strong power of masked autoencoder (MAE)~\cite{mae} as a scalable visual learner. We extensively explore the supervised and self-supervised pre-training strategies for our MixViT, and also try to implement a new pre-training strategy TrackMAE solely on the tracking datasets without ImageNet. Our TrackMAE achieves very competitive performance to the large-scale ImageNet pre-training.
We perform extensive experiments on several benchmarks and demonstrate that a series of instantiated MixFormer trackers (including MixCvT, MixCvT-L, MixViT and MixViT-L) largely surpass the state-of-the-art trackers. Especially, with plain ViT pre-trained by MAE, our MixViT achieves the AUC of 69.6\% on LaSOT and AO of 72.5\% on GOT-10k with a real-time running speed of 75 FPS.
The main contributions are summarized as follows:
\begin{itemize}
\item We propose a compact end-to-end tracking framework, termed as \emph{MixFormer}, based on Mixed Attention Modules (MAM). MAM allows for extracting target-specific features and extensive communication between target and search simultaneously.
\item We instantiate two types of MixFormer trackers, a hierarchical tracker with progressive downsampling and depth-wise convolutional projections (coined as MixCvT), and a non-hierarchical tracker built upon plain backbone of ViT (coined as MixViT). Especially, we design a pyramidal corner head for plain MixViT, which can supplement the multi-scale information for accurate target localization.
\item We empirically study a series of pre-training strategies for our MixFormer framework and uncover the important ingredients of developing an efficient and effective tracker. In particular, we come up with simple self-supervised pre-training strategy TrackMAE, which achieves competitive performance to the large-scale ImageNet pre-training.
\item For online template update, we devise a customized asymmetric attention in MAM for high efficiency, and propose an effective score prediction module to select high-quality templates, leading to an efficient and effective online transformer-based tracker. 
\item The proposed MixCvT and MixViT significantly outperform the state-of-the-art trackers on seven challenging benchmarks, including VOT2020~\cite{vot2020}, LaSOT~\cite{lasot}, TrackingNet~\cite{trackingnet}, GOT-10k~\cite{got10k}, VOT2022, UAV123~\cite{uav123} and OTB100~\cite{otb}.
\end{itemize}

A preliminary version of this work~\cite{mixformer} was published on the conference of IEEE CVPR 2022 as an oral presentation. In this new version, we make the following extensions: \romannumeral1) we instantiate two types of MixFormer framework, the original MixCvT based on Wide Mixed Attention Module and the new MixViT based on Slimming Mixed Attention Module (S-MAM). MixViT generally enjoys better tracking performance and higher running speed.
\romannumeral2) For the non-hierarchical tracker MixViT, we design a customized pyramidal corner head, supplementing some multi-scale information for bounding box localization.
\romannumeral3) We further study a series of pre-training methods for MixFormer framework and uncover the key ingredient of training a more effective tracker. In addition, we proposed a new pre-training strategy of TrackMAE solely on the tracking dataset and demonstrate its competitive performance to the large-scale ImageNet pre-training.
\romannumeral4) We provide more experimental analysis on the design of MixFormer trackers and the pre-training methods. Our MixViT-L obtains consistent performance improvement over the original MixCvT, and sets a new state-of-the-art performance on most of the tracking datasets. We also report the performance of MixViT on the latest VOT 2022 challenge and it (MixFormerL) obtains the best performance of 0.602 EAO score on the VOT-STb2022 public challenge.

\section{Related Work}
\subsection{Tracking Paradigm}
Current prevailing tracking methods often employ a three-stage architecture, containing (\romannumeral1) a backbone to extract generic features, (\romannumeral2) an integration module to fuse the information of target and search region, (\romannumeral3) classification and localization heads to produce the target states. Generally, most modern trackers~\cite{siamrpnPlus, atom,dimp,transt,siamrcnn} used ResNets as the backbones. For the most important integration module, researchers have explored various types of methods. 
First, Siamese-based trackers~\cite{siamfc,siamrpn,siamcar,dasiamrpn,siamfc++,siammask} attract a lot of attention due to its 
simplicity and efficiency.
These methods combined a correlation operation with the Siamese network, modeling the appearance similarity and correlation between the target and search. 
SiamFC~\cite{siamfc} employed a Siamese network to measure the similarity between the template and the search area with a high tracking speed. SiamRPN++~\cite{siamrpnPlus} improved cross correlation to depth-wise cross correlation, which can increase both the performance and efficiency. For more precise regression, CGACD~\cite{CGACD} and Alpha-Refine~\cite{alpha-refine} proposed pixel-wise cross correlation. PG-Net~\cite{pgnet} then replaced the cross-correlation with pixel-to-global matching correlation, which combined the advantages of the both two and achieved surprising performance.

Second, in contrast to Siamese trackers, another family of trackers~\cite{eco,kcf,bacf,strcf,dcf_,atom,dimp,fcot} learned an online target-dependent discriminative model for object tracking. 
\cite{mosse,kcf,dcf_} employed online correlation filter to distinguish targets from background. 
However, these approaches rely on complicated online learning procedures that cannot be easily deployed in an end-to-end learning architecture.
To gain benefits from end-to-end training, CFNet~\cite{cfnet} integrated the correlation filter into deep network.
DiMP~\cite{dimp} further introduced a target model predictor to online optimizing the target model instructed by the discriminative loss, which achieved leading performance in various benchmarks at that time. 

Finally, some recent trackers~\cite{transt,stark,treg,stmtrack,tmt} introduced a transformer-based integration module to capture the global dependencies between target and search area. 
CGCAD~\cite{CGACD} and SiamAttn~\cite{siamattn} introduced a correlation-guided attention and self-attention to perform discriminative tracking.
TransT~\cite{transt} developed a ego-context augment module based on self-attention and a cross-feature augment module based on cross-attention, which established associations between distant features and thus achieving excellent performance.
STARK~\cite{stark} implemented end-to-end tracking with the designed transformer encoder and decoder. However, these transformer trackers still follow the traditional paradigm of \emph{Backbone-Integration-Head}.
TREG~\cite{treg} proposed a target-aware transformer for regression branch and can generate accurate prediction in VOT2021~\cite{vot2021} challenge. Inspired by TREG, we formulate mixed attention mechanism by using both self attention and cross attention.
In this way, our MixFormer unifies the two processes of feature extraction and information integration with an iterative MAM based backbone, leading to a more compact, neat and effective end-to-end tracker. 
After our CVPR 2022 conference version~\cite{mixformer}, there are some co-occurent works in ECCV 2022 such as OSTrack~\cite{ostrack} and SimTrack~\cite{simtrack}, which also use Transformer as backbone to perform both feature extraction and information fusion. Our work starts earlier and obtains better performance than them due to our customized design of localization head, well-explored pre-trained strategies and backbone, and effective online templates selection method. 
Compared to OSTrack, our MixFormer framework works without any post-processing such as hanning window penalty, thus yielding a more simple tracker. 

\subsection{Transformers in Computer Vision}
The Vision Transformer (ViT)~\cite{vit} presented a pure transformer architecture for image modeling, with a patch embedding to convert the 2D image into a token sequence and stacked layers of multi-head self-attention (MHSA)~\cite{transformer} on the visual tokens. ViT has obtained an impressive performance on image classification and also shown the great potential on vision tasks such as object detection~\cite{vitdet}. 
Some works~\cite{swin,t2t,localvit,pvt} introduced the customized designs into ViT to capture local context in images. 
For example, PVT~\cite{pvt} incorporated a multi-stage design (without convolutions) for Transformer similar to multi-scales in CNNs.
Swin Transformer~\cite{swin} proposed a hierarchical architecture with the designed shifted window scheme.
CvT~\cite{cvt} combined CNNs and Transformers to model both local and global dependencies for image classification in an efficient way. 
Our MixFormer aims to extend vision transformer to the object tracking area.
Inspired by CvT~\cite{cvt} and ViT~\cite{vit}, we instantiate two types of MixFormer, including the \emph{hierarchical} MixCvT and the \emph{non-hierarchical} MixViT. 
MixCvT aims to utilize the inductive bias of the multi-stage design and the adoption of depth-wise convolutional projections to relieve the data requirement. MixViT try to enjoy the design simplicity and high efficiency of ViT backbone, and present a faster tracker. But due to its plain structure without inductive bias, MiXViT often requires to careful initialization, such as supervised pre-training or MAE pre-training. 

Compared to ViT~\cite{vit} and CvT~\cite{cvt}, there are some fundamental differences. (\romannumeral1) The proposed MAM performs dual attentions for both feature extraction and information integration, while ViT and CvT uses self attention to solely extract features. (\romannumeral2) The tasks are different, and the corresponding input and the head are different. We use multiple templates together with the search region as input and employ a corner-based or query-based localization head for bounding box generation, while ViT and CvT are originally designed for image classification. (\romannumeral3) We further introduce an asymmetric mixed attention and a score prediction module for the specific task of online tracking.

\subsection{Pre-training of Trackers}
Pre-training has been well recognized as an effective initialization method to improve the training stability and boost the performance of downstream tasks.
In the past few years, the most common way~\cite{detr,tsn,siamrpnPlus,atom} is to pre-train the backbone of trackers with the image classification models from ImageNet~\cite{imagenet}, containing more than one million labeled images. 
Recently, self-supervised learning~\cite{he2020momentum,chen2020simple,mae,tong2022videomae} has attracted broad attention due to its superiority to traditional supervised learning on multiple tasks. BEiT~\cite{beit} first introduced masked image modeling (MIM) via a two-stage training process, i.e. tokenizer training and token-based masked image modeling. MAE~\cite{mae} proposed an end-to-end masked autoencoder to encode representation for unmasked patches and predicted the masked patches. Concurrently, SimMIM~\cite{simmim} presented a similar masked image modeling architecture. 

In tracking area, most of the prevailing trackers pre-train the backbone on ImageNet-1k~\cite{imagenet}. OSTrack~\cite{ostrack} first employed MAE pre-training to its ViT-based tracker. In this work, we first demonstrate that the MixCvT pre-trained on a larger dataset ImageNet-22k significantly outperforms that on ImageNet-1k. Then we experiment on MixViT and uncover the differences between the MAE pre-training and supervised pre-training with detailed ablations on depth of ViT backbone. 
To further verify the effectiveness of MAE pre-training on tracking, we propose a new masked pre-training technique (TrackMAE) solely on the tracking datasets instead of the large-scale ImageNet, and demonstrate its excellent performance on MixViT pre-training.

\section{Method}
In this section, we present our end-to-end tracking framework, termed as MixFormer, based on iterative mixed attention modules (MAM), and the overall pipeline is depicted in Fig.~\ref{fig:motivation} (b). We instantiate two types of MixFormers, including the hierarchical MixCvT built upon W-MAM and the non-hierarchical MixViT built upon S-MAM. Specifically, MixCvT aims to utilize the visual inductive bias with progressive downsampling to embrace the success of hybrid structure of CNN and transform. MixViT focuses on designing a simpler and more efficient tracker with less inductive bias, but requires careful pre-training to achieve superior performance.
In Section~\ref{sec:MAM}, we introduce our proposed MAM to unify the process of feature extraction and target information incorporation. This simultaneous processing scheme will enable our feature extraction to be more specific to the corresponding tracking target. In addition, it also allows the target information integration to be performed more extensively and thus to better capture the correlation between target and search area. 
Then, we describe MixCvT and MixViT in Section~\ref{sec:mixformer} and Section~\ref{sec:mixformer_vit} respectively.
Furthermore, we introduce the details of pre-training methods in Section~\ref{sec:pretrain}.
Finally, in Section~\ref{sec:train_inf}, we describe the training and inference of MixCvT and MixViT, and the confidence score based target template update mechanism to handle object deformation in tracking procedure.

\subsection{Mixed Attention Module (MAM)}
\label{sec:MAM}

\begin{figure*}[pt]
\centering
\includegraphics[width=\linewidth]{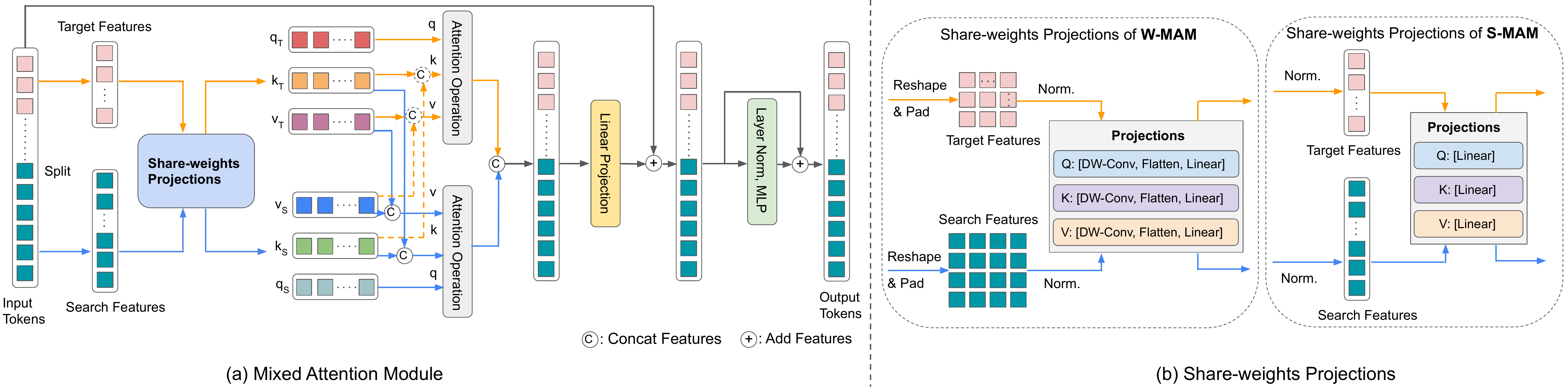}
\caption{(a)~{\bf Mixed Attention Module (MAM)} is a flexible attention operation that unifies the process of feature extraction and information integration for target template and search area. This mixed attention has dual attention operations where self-attention is performed to extract features from itself while cross-attention is conducted to communicate between target and search. This MAM could be easily implemented with a concatenated token sequence. To further improve efficiency, we propose an asymmetric MAM by pruning the target-to-search cross attention (denoted by dashed lines) (b) We instantiate two types of MAMs, including {\bf Wide MAM} for MixCvT and {\bf Slimming MAM} for MixViT. The difference only lies in the share-weights projections.
}
\label{fig_mam}
\end{figure*}

Mixed attention module (MAM) is the core design to pursue a compact end-to-end visual object tracker. The input to our MAM is the {\em target template} and {\em search area}. MAM aims to simultaneously extract their own visual features and fuse the interaction information between them.
In contrast to the original Multi Head Attention~\cite{transformer}, MAM performs dual mixed attention operations on two separate tokens sequences of target template and search area. It carries out self-attention on tokens from each sequence themselves to capture the target or search area information. Meanwhile, it conducts cross-attention between tokens from two sequences to allow information communication between target template and search area. As shown in Fig.~\ref{fig_mam}, this mixed attention mechanism could be implemented efficiently via a concatenated token sequence. In this work, we instantiate two types of MAM, including Wide MAM for MixCvT and Slimming MAM for MixViT. 
The former one introduces the desirable properties of CNN (i.e. shift, scale, and distortion invariance) to transformer architecture via the depth-wise convolutional projections. This enables the tracker to realize promising performance when the scale of training data is limited.
While the latter one pursue a ~\textit{simple} and~\textit{universal} transformer operation. The details of W-MAM and S-MAM are described as follows.

\noindent\textbf{Wide Mixed Attention Module (W-MAM).}
Formally, given a concatenated tokens of multiple targets and search, we first split it into two parts and reshape them to 2D feature maps. In order to achieve additional modeling of local spatial context, a separable depth-wise convolutional projection layer is performed on each feature map (i.e., \emph{query}, \emph{key} and \emph{value}). It also provides efficiency benefits by allowing the down-sampling in \emph{key} and \emph{value} matrices. Then each feature map of target and search is flattened and processed by a linear projection to produce queries, keys and values of the attention operation. We use $q_t$, $k_t$ and $v_t$ to represent target, $q_s$, $k_s$ and $v_s$ to represent search region. The mixed attention is defined as: 
\begin{equation}
\begin{aligned}
    & k_m = {\rm Concat}(k_t, k_s), \ \ \ v_m = {\rm Concat}(v_t, v_s), \\ 
    & {\rm Attention_{t}} = {\rm Softmax}(\frac{q_{t}k_{m}^{T}}{\sqrt{d}})v_{m},\\
    & {\rm Attention}_{s} = {\rm Softmax}(\frac{q_{s}k_{m}^{T}}{\sqrt{d}})v_{m},
\end{aligned}
\end{equation}
where $d$ represents the dimension of the key, ${\rm Attention}_{t}$ and ${\rm Attention}_{s}$ are the attention maps of the target and search respectively. It contains both self attention and cross attention which unifies the feature extraction and information integration. Then the targets token and search token are concatenated and processed by a linear projection. Finally, the concatenated token sequence is processed by a Layer Normalization and a MLP function as in Fig.~\ref{fig_mam}.

\noindent \textbf{Slimming Mixed Attention Module (S-MAM).}
While the W-MAM produces a strong visual representation through introducing translation-equivariant priors to vanilla transformer (i.e., depth-wise convolutional projection), it also possesses the a few issues. 
Primarily, it leads to a lower tracking speed, since the reshaping operation and depth-wise convolution operations for each attention element (i.e., \textit{query}, \textit{key} and \textit{value}). 
In addition, it lacks the flexibility to adapt to the recent ViT development, such as the simple masked pre-training~\cite{mae}.
To address these issues, we further present the Slimming Mixed Attention Module (S-MAM) by removing the depth-wise convolutional projection in W-MAM. The detailed structure of S-MAM is depicted in Fig.~\ref{fig_mam}.

First, we split the input token sequence into multiple targets token sequences, including the static one and online ones if we use multiple online templates, and a search token sequence. Then the split features of targets and search are processed by a layer normalization and simple linear projections. Next, the mixed attention operation is performed on the generated queries, keys and values. The remaining operations remain the same as W-MAM.

\begin{figure*}[pt]
\centering
\includegraphics[width=\linewidth]{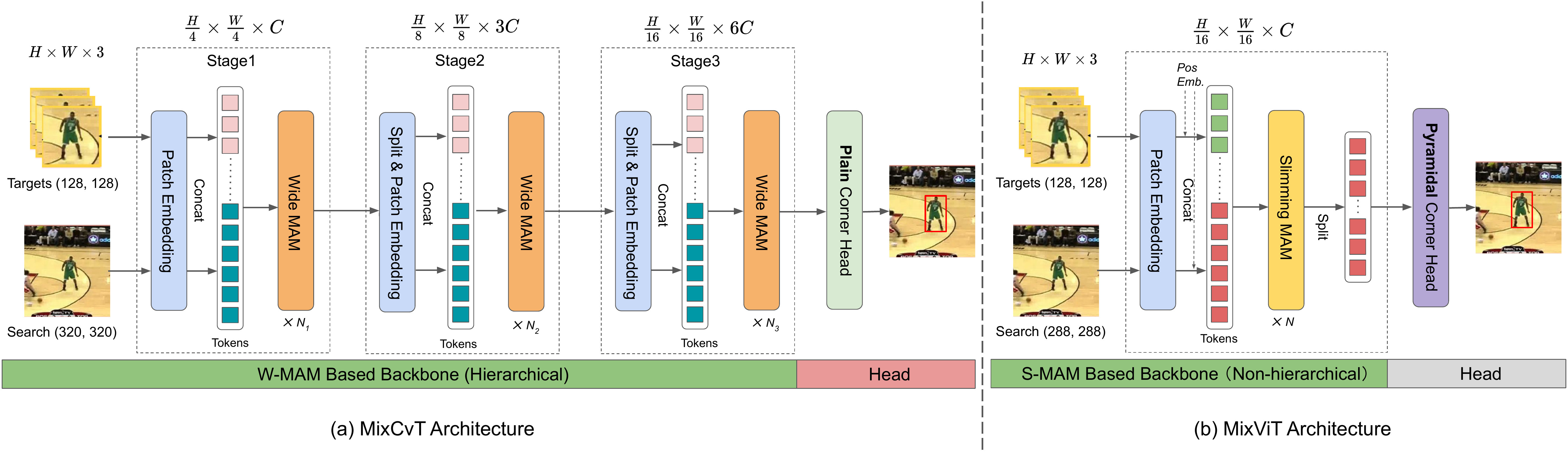}
% \vspace{-2mm}
\caption{{\bf MixFormer Tracking Pipeline.} Based on the above MAM, our MixFormer presents a compact end-to-end framework for tracking with unified process of feature extraction and target information integration. Specifically, we  instantiate two types MixFormer trackers. The first one is a hierarchical tracker with progressive downsampling based on the CvT backbone (termed as {\bf MixCvT}). The second one is a non-hierarchical tracker with plain backbone of ViT (termed as {\bf MixViT}). For these two types of trackers, we design two customized localization heads for bounding box estimation.}
\label{fig_arch}
% \vspace{-6mm}
\end{figure*}

\noindent \textbf{Asymmetric Mixed Attention Scheme.}
Intuitively, the cross attention from the targets query to search area is not so important and might bring negative influence due to potential distractors. To reduce computational cost of MAM and thereby allowing for efficiently using multiple templates to deal with object deformation, we further present a customized {\em asymmetric} mixed attention scheme by pruning the unnecessary target-to-search area cross-attention. This asymmetric mixed attention is defined as follows:
\begin{equation}
\begin{aligned}
    & {\rm Attention_{t}} = {\rm Softmax}(\frac{q_{t}k_{t}^{T}}{\sqrt{d}})v_{t}, \\
    & {\rm Attention}_{s} = {\rm Softmax}(\frac{q_{s}k_{m}^{T}}{\sqrt{d}})v_{m}.
\end{aligned}
\end{equation}
In this manner, the template tokens in each MAM could remain unchanged during tracking process and so it needs to be processed only once.

\noindent \textbf{Discussions.}
To better expound the insight of the mixed attention, we make a comparison with the attention mechanism used by other transformer trackers.
Different with our mixed attention, TransT~\cite{tt} uses ego-context augment and cross-feature augment modules to perform self attention and cross attention progressively \emph{in two steps}. Compared to the transformer encoder of STARK~\cite{stark}, our MAM shares a similar attention mechanism but with three notable differences. First, in W-MAM, we incorporate the spatial structure information with a depth-wise convolution while they use positional encoding. More importantly, our MAM is built as a multi-stage backbone for both joint feature extraction and information integration, while they depend on a separate CNN backbone for feature extraction and only focus on information integration in another single stage. Finally, we also propose a different asymmetric MAM to further improve the tracking efficiency without much accuracy drop.

\subsection{Hierarchical MixFormer: MixCvT}
\label{sec:mixformer}

\noindent \textbf{Overall Architecture.}
Based on the W-MAM blocks, we build the hierarchical MixFormer tracker of MixCvT. The main idea of MixCvT is to progressively extract coupled features for target template and search area, and deeply perform the information integration between them. Basically, it comprises two components: a backbone composed of iterative W-MAMs, and a simple localization head to produce the target bounding box.
Compared with other prevailing trackers by decoupling the steps of feature extraction and information integration, it leads to a more compact and neat tracking pipeline only with a single backbone and tracking head, without an explicit integration module or any post-processing. 
The overall architecture is depicted in Fig.~\ref{fig_arch}. 

\noindent \textbf{W-MAM Based Backbone.}
Our goal is to couple both the generic feature extraction and target information integration within a unified transformer-based architecture. The W-MAM based backbone employs a progressive multi-stage architecture design. Each stage is composed of $N$ MAM layers operating on the same-scaled feature maps with the identical channel number. All stages share the similar architecture, which consists of an overlapped patch embedding layer and $N_i$ wide mixed attention modules. 

Specifically, given $T$ templates (i.e., the first template and $T-1$ online templates) with the size of $T\times H_{t}\times{W_{t}}\times3$ and a search region (a cropped region according to the previous target states) with the size of $H_{s}\times{W_{s}}\times3$, we first map them into overlapped patch embeddings using a \emph{convolutional Token Embedding} layer with stride $4$ and kernel size $7$. 
The convolutional token embedding layer is introduced in each stage to grow the channel resolution while reducing the spatial resolution. 
Then we flatten the patch embeddings and concatenate them, yielding a fused token sequence with the size of $(T\times\frac{H_{t}}{4}\times\frac{W_t}{4}+\frac{H_s}{4}\times\frac{W_s}{4})\times C$, where $C$ equals to 64 or 192, $H_t$ and $W_t$ is 128, $H_s$ and $W_s$ is 320 in this work. 
After that, the concatenated tokens pass through $N_i$ target-search MAM to perform both feature extraction and target information incorporation. 
Finally, we obtain the token sequence of size $(T\times\frac{H_{t}}{16}\times\frac{W_t}{16}+\frac{H_s}{16}\times\frac{W_s}{16})\times6C$. More details about the MAM backbones could be found in the Section~\ref{model_arch} and Table~\ref{tab:arch}.
Before passed to the prediction head, the search tokens are split and reshaped to the size of $\frac{H_{s}}{16}\times\frac{W_s}{16}\times6C$. 
Particularly, we do not apply the multi-scale feature aggregation strategy, commonly used in other trackers (e.g., SiamRPN++~\cite{siamrpnPlus}).

\noindent \textbf{Corner Based Localization Head.}\label{corner_head}
Inspired by the corner detection head in STARK~\cite{stark}, we employ a fully-convolutional corner based localization head to directly estimate the bounding box of tracked object, solely with several $\rm Conv$-$\rm BN$-$\rm ReLU$ layers for the top-left and the bottom-right corners prediction respectively. Finally, we can obtain the bounding box by computing the expectation over corner probability distribution~\cite{GFloss}. The difference with STARK lies in that ours is a fully convolutional head while STARK highly relies on both encoder and the decoder with more complicated design.
% \vspace{-3mm}

\noindent \textbf{Query Based Localization Head.}\label{query_head}
Inspired by DETR~\cite{detr}, we propose to employ a simple query based localization head. This sparse localization head can verify the generalization ability of our MAM backbone and yield a pure transformer-based tracking framework. Specifically, we add an extra learnable \emph{regression token} to the sequence of the final stage and use this token as an anchor to aggregate information from entire target and search area. Finally, a FFN of three fully connected layers is employed to directly regress the bounding box coordinates. This framework does not use any post-processing technique either.

\begin{figure*}[t]
\centering
\includegraphics[width=\linewidth]{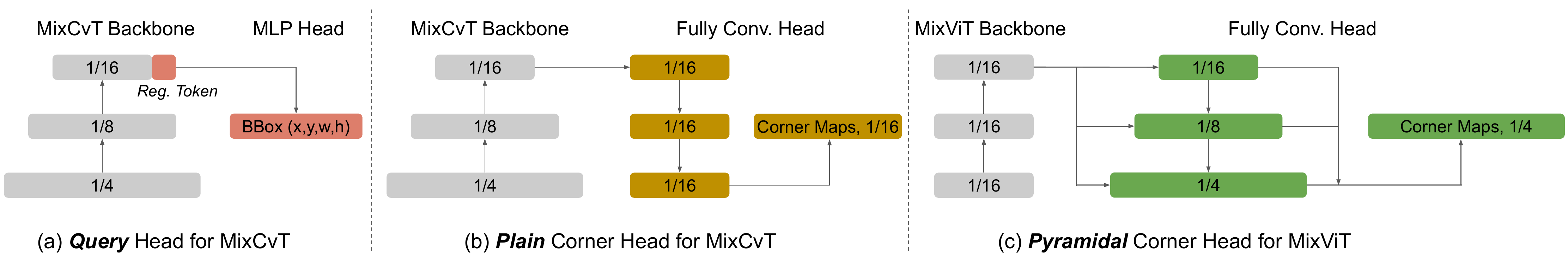}
% \vspace{-6mm}
\caption{{\bf Different Heads} employed in MixCvT and MixViT. Two types of head are developed to validate the generalization of MixFormer. We also explore building a pyramidal corner head for the non-hierarchical MixViT to supplement the multi-scale information.}
\label{fig_heads}
% \vspace{-6mm}
\end{figure*}

\subsection{Plain MixFormer: MixViT}
\label{sec:mixformer_vit}

\noindent \textbf{Overall Architecture.}
To remove the designs of the hierarchical structure and the convolutional projections, we build a simple and non-hierarchical tracker, termed as MixViT. The architecture of MixViT is depicted in Fig.~\ref{fig_arch} (b). MixViT is composed of a patch embedding layer, several Slimming Mixed Attention Module (S-MAM) on the token sequence and a new pyramidal corner head on top. 
In this section, we describe the Slimming Mixed Attention Module, S-MAM based backbone, different positional embeddings for MixFormer-ViT and the proposed pyramidal corner head.

\noindent \textbf{S-MAM Based Backbone.}
\label{sec_smam_backbone}
Compared to MixCvT tracker, the following adaptions are made in MixViT tracker. 
First, the backbone of MixViT is non-hierarchical and one-stage, composed of a single non-overlapped patch embedding layer and multiple slimming mixed attention modules. This avoids a tremendous number of tokens in the high-resolution stages as in MixCvT, thus guaranteeing a high tracking efficiency.
Second, we explore different types of positional embeddings for targets and search tokens to explicitly encode the absolute position information.
Third, since MixViT backbone has the same trainable-weights with vanilla Vision Transformer~\cite{vit}, it can be compatible with the recent ViT development, such as MAE pretraining,.

First, given $T$ templates with the size of $T\times H_{t} \times W_{t} \times 3$ and the search region with the size of $H_{s} \times W_{s} \times 3$, we map them into non-overlapped patch embeddings using a convolutional token embedding layer with kernel size of 16 and stride of 16.
Then we add two different positional embeddings to search region and templates patch embeddings respectively. Three different types of positional embeddings are experimented as in Section~\ref{sec_pos_emb}. Next, we concatenate the templates (with size of $T \times \frac{H_{t}}{16} \times \frac{W_{t}}{16} \times C$) and search region tokens (with size of $\frac{H_{s}}{16} \times \frac{W_{s}}{16} \times C$) and input it to multiple Slimming Mixed Attention Modules. Finally, the split search region token is prepared to input the localization head.

\noindent \textbf{Positional Embedding.}
\label{sec_pos_emb}
We investigate three types of positional embeddings in this work to give a thorough inspection to the targets and search region position modeling.
For all of the three types, we use two positional embeddings of different length for targets and search region. The differences between the three types of positional embeddings lie in the initialization methods and whether to be learnable. 
For the first type, we adopt 2D bilinear interpolation to interpolate the pretrained positional embeddings from MAE-pretrained ViT models. And the embeddings are set to be frozen during training process.
For the second one, the same initialization method is employed as the first type. However, the embeddings are set to be learnable so as to realize dynamic adjustment.
For the last one, we use frozen sin-cos positional embeddings as adopted in vanilla ViT~\cite{vit}. 

\noindent \textbf{Pyramidal Corner Head.}
\label{pyramid_head}
Since MixViT adopts a non-hierarchical backbone and may lacks the multi-scale information, we build a pyramidal corner heads from only the last feature map of MixViT backbone. The structure of pyramidal head is depcited in Fig.~\ref{fig_heads} (c).
Specifically, we first build the feature pyramid via several convoluational layers and interpolation layers, thereby producing three different-resolution feature maps. Then, the multi-scale feature maps are fused by convolutional layers to get robust representations.
We find that, compared with plain corner head employed in MixCvT, the pyramidal one can obtain better performance as shown in experiments. 

\subsection{Pre-training of MixFormer}
\label{sec:pretrain}
In this work, we find that the pre-training of backbone plays an important role in building an effective MixFormer tracker. Thanks to the flexibility of attention operation, the pre-trained CvT or ViT models on ImageNet for image classification could be directly adapted to initialize the weights of backbone of our MixFormer trackers. Specifically, we investigate different supervised pre-trained models for MixCvT, and both supervised and self-supervised pre-trained models for MixViT.
Inspired by the excellent performance of masked pre-training~\cite{mae}, we further design a masked auto-encoder on the commonly used tracking datasets (i.e., LaSOT~\cite{lasot}, TrackingNet~\cite{trackingnet}, GOT-10k~\cite{got10k} and COCO~\cite{coco}), and apply it as the pretraining weights for MixViT.

\noindent {\bf Supervised Pre-training on ImageNet.}
For MixCvT, the model weights of CvT trained on ImageNet-1k or ImageNet-22k are used as the initialization of MixCvT backbone. 
Pre-training on ImageNet-22k provides stronger visual representations than ImageNet-1k.
For MixViT, we carry out the supervised pre-training via the recent DeiT3~\cite{deit3} model, which successfully trains vanilla ViT models with high performance on the ImageNet-1k and ImageNet-21k. It is noteworthy that we add LayerScale in transformer blocks of MixViT when pretraining it with DeiT3 to keep consistency. 

\noindent {\bf Self-supervised Pre-training on ImageNet.}
Inspired by the great success of masked autoencoder (MAE) pre-training~\cite{mae} and its application in object detection~\cite{vitdet}, we also investigate the MAE pre-training in our MixViT.
To better investigate the difference in the learned representations from supervised and self-supervised pre-training, we perform detailed studies on the effect of pre-training by varying the ViT depth. We observe that the different behaviors of supervised pre-training and MAE pre-training, where supervised pre-trained models tend to learn high-level semantic representations in the latter stages, while the MAE pre-trained models seem to focus on learning the low-level signal structure progressively along all layers. Therefore, reducing several layers in the supervised pre-trained model leads to very small performance drop as the high-level semantic features in the latter stages contribute little to object tracking. In addition, we further increase the depth of the MAE pre-trained ViT by borrowing transformer blocks from the MAE decoder, and obtain the further performance improvement.

To combine the advantages of masked pre-training and visual inductive prior of progressive modeling, we further explore the ConvMAE~\cite{convmae} pretraining in our MixViT tracker.
The architecture of ConvMAE pre-trained model is similar to the plain ViT, except that it replaces the original one-layer strided patch embedding with a lightweight CNN for progressive patch embedding. So, we can easily adapt the ConvMAE pre-trained model to our MixViT by simply changing the patch embedding layer to the corresponding CNN.
The modified MixViT with ConvMAE pre-training consists of a three-stage patch embedding CNN, the plain ViT encoder of S-MAM, and a pyramidal corner head. The patch embedding CNN uses simple convolution blocks to separately transform the search and templates to token embeddings. The details on ViT backbone design is similar to the original MAE pretraining as shown in Table~\ref{tab:arch_mixvit}.

\noindent {\bf Simple MAE Pre-training on Tracking Datasets.}
Inspired by the excellent performance MAE pre-training~\cite{mae} and its data-efficiency~\cite{tong2022videomae} on the training samples, we are wondering whether we are able to pre-train our MixVit with MAE solely on the common tracking datasets without using the large-scale ImageNet dataset. First, we design a simple MAE baseline by directly pre-training a masked auto-encoder on the tracking datasets and use this pre-trained MAE to initialize our MixViT.
In details, we first train a masked auto-encoder on tracking datasets (including LaSOT, TrackingNet, GOT-10k and coco) and then fine-tune MixViT on the same tracking datasets. 
The main procedure of training masked auto-encoder keeps consistent with MAE~\cite{mae} excepting for the datasets and data augmentations.
Surprisingly, we observe that MixViT with self-supervised pretraining solely on the tracking datasets outperforms that with training from scratch by a large margin.

\begin{figure}[t]
\centering
\includegraphics[width=\linewidth]{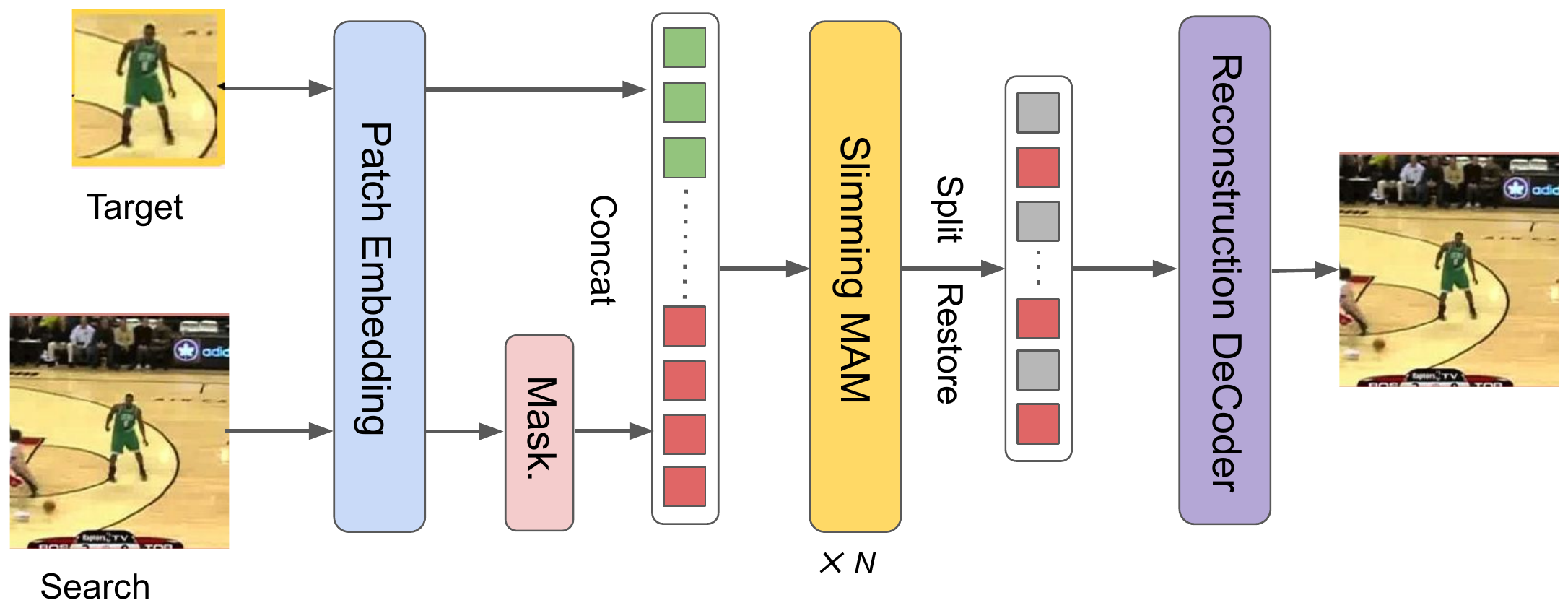}
\vspace{-3mm}
\caption{Framework of \textbf{TrackMAE}. We perform random masking on the search image with a mask ratio of 75\% while keeping the target template unmasked. The decoder reconstructs the invisible patches in the search image.}
\vspace{-3mm}
\label{fig_trackmae}
\end{figure}

\noindent {\bf TrackMAE Pre-training on Tracking Datasets.}
To better adapt the MAE pre-training to our MixViT tracker, we present a new self-supervised pre-training method for visual object tracking, termed as  {\bf TrackMAE}, as shown in Fig.~\ref{fig_trackmae}.
TrackMAE inherits the simple pipeline of random token dropping and reconstructing the masked ones. 
However, to make it more suitable for the tracking framework and narrow the gap between tracker pre-training and fine-tuning, the search images are masked with a mask ratio of 75\% while the templates are fully visible and are placed for information integration.
The decoder reconstructs the invisible parts of search images.
Especially, we only use the tracking datasets for TrackMAE pre-training.

\subsection{Training and Inference}
\label{sec:train_inf}

\noindent {\bf Training.}
The training process of our MixFormer generally follows the standard training recipe of current trackers~\cite{stark,transt}. We first pre-train our MixCvT and MixViT backbones with CvT model~\cite{cvt} and encoders of MAE~\cite{mae} respectively, and then fine-tune the whole tracking framework on the tracking datasets. 
For MixCvT training, a combination of ${L_1}$ loss and GIoU loss~\cite{giou} is employed as follows:
\begin{equation}
    L_{loc} = \lambda_{L1} L_1(B_i, \hat{B_i}) + \lambda_{giou} L_{giou}(B_i, \hat{B_i}),
\end{equation}
where $\lambda_{L1}=5$ and $\lambda_{giou}=2$ are the weights of the two losses, $B_i$ is the ground-truth bounding box and $\hat{B_i}$ is the predicted bounding box of the targets. 
For MixViT training, we replace the GIoU loss with the CIoU loss~\cite{diou}.

\begin{figure}[t]
\centering
\includegraphics[width=\linewidth]{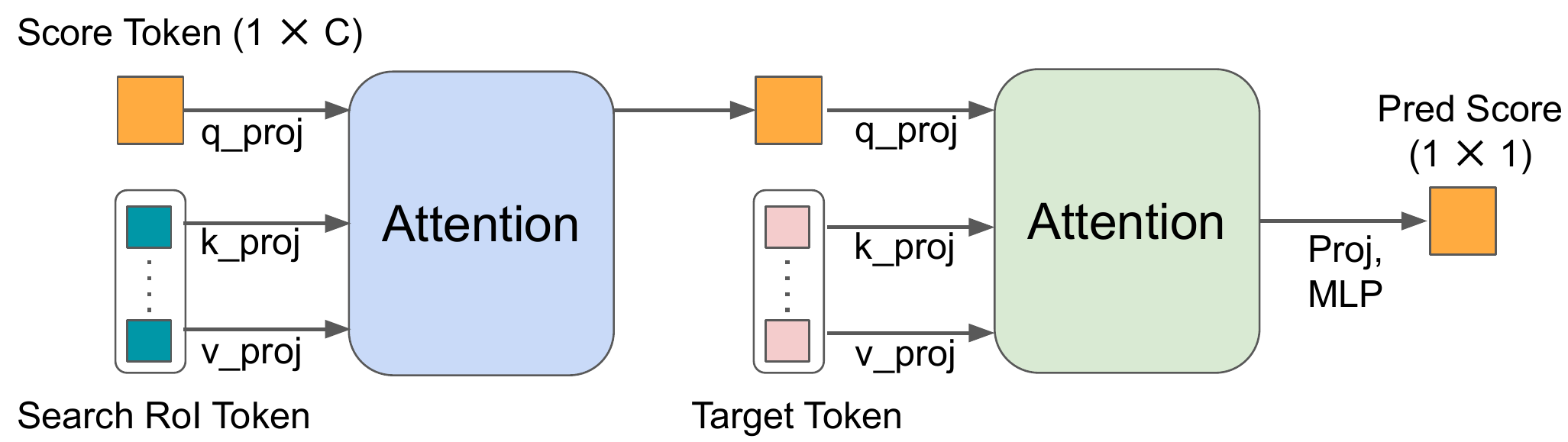}
% \vspace{-6mm}
\caption{Structure of the \textbf{Score Prediction Module} (SPM). 
}
% \vspace{-5mm}
\label{fig_score_module}
\end{figure}

\noindent {\bf Template Online Update.}
Online templates play an important role in capturing temporal information and dealing with object deformation and appearance variations. 
However, poor-quality online templates may lead to tracking performance degradation. As a consequence, we introduce a \textbf {Score Prediction Module} (SPM), described in Fig.~\ref{fig_score_module}, to select reliable online templates determined by the predicted confidence score. The SPM is composed of two attention blocks and a three-layer perceptron. First, a learnable \emph{score token} serves as a query to attend the search ROI tokens. It enables the score token to encode the mined target information. Next, the score token attends to all positions of the initial target token to implicitly compare the mined target with the first target. Finally, the score is produced by the MLP layer and a sigmoid activation. The online template is treated as negative when its predicted score is below than 0.5.
For the SPM training, it is performed after the backbone training and we use a standard cross-entropy loss:
\begin{equation}
\begin{split}
L_{score} = y_i{\rm log}(p_i)+(1-y_i){\rm log}(1-p_i), 
\end{split}
\end{equation}
where $y_i$ is the ground-truth label and $p_i$ is the predicted confidence score.

\noindent \textbf{Discussions on SPM.}
We compare our SPM with the score prediction branch in STARK~\cite{stark}.
In contrast to STARK, \romannumeral1) we explicitly input the tracked object RoI feature (cropped based on MixFormer’s tracking results) of the current frame to the SPM, while STARK implicitly extract the target information through an encoder-decoder structure;
\romannumeral2) Unlike STARK, there is no decoder in MixFormer backbone, so that we introduce an extra score token and two attention modules.
\romannumeral3) Our SPM is composed of two progressive attention blocks to take comparison between tracked object and the target, and a MLP head to produce the online template confidence score. In contrast, STARK only employs a three-layer perceptron head to output the template confidence score.

\noindent {\bf Inference.} During inference, multiple templates, including one static template and $N$ dynamic online templates, together with the cropped search region are fed into MixFormer trackers to produce the target bounding box and the confidence score. We update the online templates only when the update interval is reached and select the sample with the highest confidence score. 
Especially, the predicted bounding box is served as the final output without any post-processing such as cosine window or bounding box smoothing.

\begin{table}[pt]
\centering
\caption{W-MAM based backbone architectures for MixCvT and MixCvT-L. The input is a tuple of templates with shape of $128\times128\times3$ and search region with shape of $320\times320\times3$. $S$ and $T$ represent for the search region and template. $H_i$ and $D_i$ is the head number and embedding feature dimension in the $i$-th stage. $R_i$ is the feature dimension expansion ratio in the MLP layer. 
}
\fontsize{7pt}{3.5mm}\selectfont
\setlength{\tabcolsep}{0.7mm}{
\resizebox{\columnwidth}{!}{
\begin{tabular}{c|c|l|c|c}
\hline
\multicolumn{1}{l|}{}     
& Output Size
& Layer Name                                                          
& MixCvT
& MixCvT-L \\ 
\hline

\multirow{3}{*}{Stage1}    
& $\begin{array}{c} S: 80\times80,\\ T: 32\times32
\end{array}$
& Conv. Embed.                                                     
& \multicolumn{1}{c|}{$7\times7$, $64$, stride $4$}                   
& $7\times7$, $192$, stride $4$               \\ 
\cline{2-5} 

& $\begin{array}{c} S: 80\times80,\\ T: 32\times32
\end{array}$
& \begin{tabular}[c]{@{}l@{}}W-MAM \end{tabular} 
&             
$\left[
\begin{array}{c}
    %  3\times3, 64\\
     H_1 = 1 \\
     D_1=64 \\
     R_1=4
\end{array}
\right] \times 1$
&      
$\begin{bmatrix}
\begin{array}{c}
    %  3\times3, 192\\
     H_1=3\\
     D_1=192\\
     R_1=4  
\end{array}
\end{bmatrix} \times 2$
\\ \hline

\multirow{3}{*}{Stage2}    
& $\begin{array}{c} S: 40\times40,\\ T: 16\times16
\end{array}$
&  Conv. Embed.
& \multicolumn{1}{c|}{$3\times3$, $192$, stride $2$}                   
& $3\times3$, $768$, stride $2$
\\ \cline{2-5} 
                           
& $\begin{array}{c} S: 40\times40,\\ T: 16\times16
\end{array}$
& \begin{tabular}[c]{@{}l@{}} W-MAM \end{tabular} 
&             
$\begin{bmatrix}
\begin{array}{c}
    %  3\times3, 192\\
     H_2=3 \\
     D_2=192 \\
     R_2=4 
\end{array}
\end{bmatrix} \times 4$
&             
$\begin{bmatrix}
\begin{array}{c}
    %  3\times3, 768\\
     H_2=12, \\ D_2=768,\\
     R_2=4  
\end{array}
\end{bmatrix} \times 2$
\\ \hline

\multirow{3}{*}{Stage3}    
& $ \begin{array}{c} S: 20\times20,\\ T: 8\times8
\end{array} $      
& Conv. Embed.
& \multicolumn{1}{c|}{$3\times3$, $384$, stride $2$}                   
& $3\times3$, $1024$, stride $2$
\\ \cline{2-5} 
                           
& $\begin{array}{c} S: 20\times20,\\ T: 8\times8
\end{array}$
& \begin{tabular}[c]{@{}l@{}}W-MAM\end{tabular} 
&             
$\begin{bmatrix}
\begin{array}{c}
    %  3\times3, 384\\
     H_3=6 \\
     D_3=384 \\
     R_3=4  
\end{array}
\end{bmatrix} \times 16$
&
$\begin{bmatrix}
\begin{array}{c}
    %  3\times3, 1024\\
     H_3=16 \\
     D_3=1024 \\
     R_3=4 
\end{array}
\end{bmatrix} \times 12$
\\ \hline

\multicolumn{3}{c|}{MACs}  
& $35.61$ M
& $183.89$ M
\\ \hline
\multicolumn{3}{c|}{FLOPs}  
& $23.04$ G
& $127.81$ G
\\ \hline
\multicolumn{3}{c|}{Speed (RTX 8000)}  
& $31$ FPS 
& $27$ FPS 
\\ \hline
\end{tabular}
}}
\label{tab:arch}
% \vspace{-4mm}
\end{table}

\begin{table}[t]
\centering
\caption{S-MAM based backbone architectures for MixViT and MixViT-L. $S$ and $T$ represent for the search region and the template respectively. `$*$' denotes for the hidden dim in the third stage of MixViT with ConvMAE pre-training.}
\vspace{-2mm}
\resizebox{\linewidth}{!}{
\begin{tabular}{l|ccccc}
\toprule
\multirow{2}{*}{Model} & \multirow{2}{*}{Layers} & Hidden
& Input Size
&  \multirow{2}{*}{Params} & Speed \\
 & & Dims & ($S$ / $T$) & & (RTX 8000) \\
\midrule 
MixViT   &   12   &        768      & ($288\times288$) / ($128\times128$) &   97M & 75 FPS \\
MixViT-L  &   24   &       1024      & ($384\times384$) / ($192\times192$) &  318M & 13 FPS \\
\midrule
MixViT (ConvMAE)   &   15   &        768$^{*}$     & ($288\times288$) / ($128\times128$) & 96  M & 45 FPS \\
MixViT-L (ConvMAE)  &   24   &       1024$^{*}$      & ($384\times384$) / ($192\times192$) & 286 M & 10 FPS \\
\bottomrule
\end{tabular}}
\vspace{-3mm}
\label{tab:arch_mixvit}
\end{table}

\section{Experiments}

\subsection{Implementation Details}
Our trackers are implemented using Python 3.6 and PyTorch 1.7.1. The MixFormer training is conducted on 8 Tesla V100 GPUs. Especially, MixFormer is a neat tracker {\bf\em without} some post-processing strategies. The inference of MixCvT and MixViT are performed on a Quadro RTX-8000 GPU.
% \vspace{-4mm}

{\noindent \bf{Architectures.}}\label{model_arch}
As shown in Table~\ref{tab:arch}, we instantiate two types of MixCvT trackers, denoted by MixCvT and MixCvT-L, with different parameters and FLOPs by varying the number of W-MAM blocks and the hidden feature dimension in each stage. The backbones of MixCvT and MixCvT-L are initialized with the CvT-21 and CvT24-W~\cite{cvt} (first 16 layers are employed) pre-trained on ImageNet~\cite{imagenet} respectively.
In addition, as shown in Table~\ref{tab:arch_mixvit}, we instantiate two models of MixViT, i.e., MixViT and MixViT-L, by varying the number of S-MAM blocks and the hidden feature dimension. The backbone weights of MixViT and MixViT-L are initialized with corresponding MAE encoders. Besides, we try ConvMAE pre-training that has comparable parameters and Flops to the plain ViT for the initialization of MixViT.

% \vspace{-4mm}

\noindent \textbf{Training.} 
The training set includes TrackingNet~\cite{trackingnet}, LaSOT~\cite{lasot}, GOT-10k~\cite{got10k} and COCO~\cite{coco} training dataset, which is the same as DiMP~\cite{dimp} and STARK~\cite{stark}.
While for GOT-10k test, we re-train our trackers by only using the GOT-10k train split following its standard protocol. 
The whole training process of MixFormer consists of two stages, which contains the first 500 epochs for the tuning of backbones and heads, and extra 40 epochs for score prediction head tuning. 
For MixCvT training, we use ADAM~\cite{adam} with weight decay of $1\times10^{-4}$. The learning rate is initialized as $1\times10^{-4}$ and decreased to $1\times10^{-5}$ at the epoch of 400. 
The sizes of search images and templates are $320\times320$ pixels and $128\times128$ pixels respectively. For data augmentations, we use the horizontal flip and brightness jittering.
For MixViT training, we use ADAMW with weight decay of $1\times10^{-4}$ and learning rate of weight decay of $4\times10^{-4}$ and decreased to $4\times10^{-5}$ at the epoch of 400.
In MixViT setting, the sizes of search images and templates are $288\times288$
pixels and $128\times128$ pixels respectively. In the MixViT-L setting, to improve the representation power of trackers, the sizes of search images and templates are set to $384\times384$ pixels and $192\times192$ pixels, respectively.

% \vspace{-4mm}

\noindent \textbf{Inference.}
We use the first template and multiple online templates together with the current search region as input of MixFormer. The dynamic templates are updated when the update interval of 200 is reached by default (the update interval varies slightly with different test datasets). The templates with the highest predicted score in the interval is selected to replace the previous one.

\subsection{Exploration Studies}
To verify the effectiveness and give a thorough analysis on our proposed MixFormer trackers (i.e. MixViT-L, MixViT, MixCvT-L and MixCvT) and pre-training methods, we perform detailed ablation studies on the large-scale LaSOT dataset. AUC is employed as the evaluation metric.

% \vspace{-4mm}
\subsubsection{Study on Mixed Attention Module}

\begin{table}[pt]
\caption{Analysis of the MAM based framework. `-' denotes the component is not used. SAM represents for self attention module, CAM for cross attention module and W-MAM for the proposed mixed attention module. The numbers in brackets represent the number of the blocks. Performance is evaluated on LaSOT.}
    \centering
    % \fontsize{7.5}{8.5}\selectfont  
    \fontsize{7.5pt}{3.5mm}\selectfont
    \setlength{\tabcolsep}{0.85mm}{
    % \resizebox{\linewidth}{!}{
    % \vspace{-1mm}
    % \resizebox{\columnwidth}{!}{
    % \small
    \begin{tabular}{c|ccc|ccc}
        \hline
        \text{\#}&Backbone&Integration&Head&Params.&FLOPs&AUC\\
        \hline
        1& SAM(21) & CAM(1) & Corner & 37.35M & 20.69G & 59.8 \\
        2& SAM(21) & CAM(3) & Corner & 40.92M & 22.20G & 60.5 \\
        3& SAM(21) & ECA+CFA(4) & Corner & 49.75M & 27.81G & 66.9 \\
        \hline
        \hline
        4& SAM(20)+W-MAM(1) & - & Corner & 35.57M & 19.97G & 65.8 \\
        5& SAM(15)+W-MAM(6) & - & Corner & 35.67M & 20.02G & 66.2 \\
        6& SAM(10)+W-MAM(11) & - & Corner & 35.77M & 20.07G & 67.4 \\
        7& SAM(5)+W-MAM(16) & - & Corner & 35.87M & 20.12G & 68.1 \\
        \hline
        \hline
        8& W-MAM(21) & - & Corner & 35.97M & 20.85G & \cellcolor{gray!20}\textbf{68.4} \\
        9& W-MAM(21) & - & Query & 31.46M & 19.13G & 66.0 \\
        \hline
    \end{tabular}
    }
    % }
    \label{tab_component}
% \vspace{-6mm}
\end{table}

\noindent \textbf{Simultaneous process vs. Separate process.} 
As the core part of our MixFormer is to unify the procedure of feature extraction and target information integration in a single module, we compare it to the separate processing architecture (e.g. TransT~\cite{tt}). The comparison results are shown in Table~\ref{tab_component} \#1, \#2, \#3 and \#8. Experiments of \#1 and \#2 are end-to-end trackers comprising a self-attention based backbone, ${n}$ cross attention modules to perform information integration and a corner head. \#3 is the tracker with CvT as backbone and TransT's ECA+CFA(4) as interaction. Experiment of \#8 is our MixCvT {\bf \textit{without}} multiple online templates and asymmetric mechanism, denoted by MixCvT-Base. MixCvT-Base largely increases the model of \#1 (using one CAM) and \#2 (using three CAMs) by 8.6\% and 7.9\% with smaller parameters and FLOPs. This demonstrates the effectiveness of unified feature extraction and information integration, as both of them would benefit each other.
% \vspace{-3mm}

\begin{table}[pt]
\caption{Ablation for asymmetric mixed attention mechanism. All models are deployed without online templates update mechanism and pretrained on ImageNet-22k. Performance is evaluated on LaSOT.}
% \small
    \centering
    \fontsize{8.5pt}{4.5mm}\selectfont
    % \fontsize{3}{4}\selectfont  
    % \setlength{\tabcolsep}{1mm}{
    \setlength{\tabcolsep}{1.2mm}{
    \begin{tabular}{c|c|cc}
        \hline
        Tracker & Asymmetric.&FPS (RTX 8000)&AUC\\
        \hline
        MixCvT & - & 25 & \cellcolor{gray!20}\textbf{68.4} \\
        MixCvT & \checkmark & 31 & 68.1 \\
        \hline
    \end{tabular}
    }
    % }
% \vspace{-1mm}
\label{tab_asym}
\end{table}
% \vspace{-4mm}

% \vspace{-2mm}
\noindent \textbf{Study on stages of MAM.}
To further verify the effectiveness of the MAMs, we conduct experiments as in Table~\ref{tab_component} \#4, \#5, \#6, \#7 and \#8, to investigate the performance of different numbers of W-MAM in our MixCvT. We compare our W-MAM with the self-attention operations (SAM) with out cross-branch information communication. We find that more W-MAMs contribute to higher AUC score.
It indicates that extensive target-aware feature extraction and hierarchical information integration play a critical role to construct an effective tracker, which is realized by the iterative MAM. Especially, when the number of W-MAM reaches 16, the performance reaches 68.1, which is comparable to the MixFormer-Base containing 21 W-MAMs.

\noindent \textbf{Study on asymmetric MAM.}
The asymmetric MAM is used to reduce computational cost and allows for usage of multiple templates during online tracking. As shown in Table~\ref{tab_asym}, the asymmetric MixCvT-Base tracker increases the running speed of 24\% while achieving a comparable performance, which demonstrates asymmetric MAM is important for building an efficient tracker.
% \vspace{-4mm}

\subsubsection{Study on MixCvT~\& MixViT Architectures}
To compare the hierarchical tracker MixCvT and non-hierarchical tracker MixViT fairly, we perform ablation studies without any pre-training (i.e., training from scratch) and with the same localization head. We aim to explore whether the MixCvT backbone exhibits superiority of the progressive downsampling and locality modeling over MixViT backbone and we gradually evolve MixViT into MixCvT as in Table~\ref{tab_arch_ana}.
For fair comparison, all models in Table~\ref{tab_arch_ana} are trained from scratch, employ plain corner head, and are deployed without online scheme. Notably, the Batch Normalization in original MixCvT are substituted to Layer Normalization, otherwise the performance is extremely poor.
The MixViT model as in the first line of Table~\ref{tab_arch_ana} obtains AUC of 56.2 without multi-stage strategies and depth-wise convolutional projections.
When extending it to a multi-stage one with three convolutional patch embeddings as MixCvT, the AUC score increases by 2.6\%. Further, when adding depth-wise convolutional projections, the AUC score slightly increases by 0.2\%. 
We can conclude that, MixCvT obtains superior performance to MixViT under the setting of training from scratch only on the tracking datasets, and the multi-stage strategy plays a more important role than depth-wise convolutional projections. It implies that incorporating visual inductive bias into backbone design has the benefits to relieve the risk of over-fitting and improve the generalization ability when without extra pre-training techniques.

\begin{table}[pt]
\caption{Study on MixCvT and MixViT backbone architectures, including the analysis of multi-stage downsampling scheme and depth-wise convolutional projections. For fair comparison, all models are trained from scratch, employ plain corner head and are deployed without online scheme.}
% \small
    \centering
    \fontsize{8.5pt}{4.5mm}\selectfont
    % \fontsize{8}{9}\selectfont  
    \setlength{\tabcolsep}{1.2mm}{
    % \resizebox{\linewidth}{!}{
    % \vspace{-1mm}
    % \resizebox{\columnwidth}{!}{
    % \small
    \begin{tabular}{l|ccc|c}
        \hline
       Tracker & Backbone & Multi-stage & DW-Conv Proj. & AUC \\
        \hline
        MixViT & S-MAM(12) & -   & - &  56.2 \\
        - & S-MAM(12) & \checkmark  & - & 58.8 \\
        MixCvT & W-MAM(12) & \checkmark  & \checkmark & \cellcolor{gray!20}\textbf{59.0} \\
        \hline
    \end{tabular}
    }
    % }
    \label{tab_arch_ana}
% \vspace{-7mm}
\end{table}

\begin{table}[pt]
\caption{Study on different heads and different positional embeddings employed in MixViT. All models are pretrained with MAE encoders and are deployed without online scheme.}
% \small
    \centering
    \fontsize{8.5pt}{4.5mm}\selectfont
    % \fontsize{8}{9}\selectfont  
    \setlength{\tabcolsep}{1.2mm}{
    % \resizebox{\linewidth}{!}{
    % \vspace{-1mm}
    % \resizebox{\columnwidth}{!}{
    % \small
    \begin{tabular}{c|cc|c}
        \hline
        Tracker & Head & Positional Emb. &AUC \\
        \hline
        MixViT & Plain Corner Head & V1 &  68.4 \\
        MixViT & Plain Corner Head & V2 & 68.4 \\
        MixViT & Plain Corner Head & V3 & \cellcolor{gray!20}\textbf{68.5} \\
        \hline
        \hline
        MixViT & Plain Corner Head & V3 & 68.5 \\
        MixViT & Pyramidal Corner Head & V3 & \cellcolor{gray!20}\textbf{69.0} \\
        \hline
    \end{tabular}
    }
    % }
   \label{tab_mixvit_heads}
% \vspace{-7mm}
\end{table}

\subsubsection{Study on Heads and Positional Embeddings}
% \vspace{-4mm}
\noindent \textbf{MixCvT's localization heads.}
To verify the generalization ability of our MAM backbone, we evaluate the MixCvT-Base with two types of localization head as described in Section~\ref{corner_head} (fully convolutional head vs. query based head). The results are shown as in Table~\ref{tab_component}, where \#8 and \#9 are for the corner head and the query-base head respectively.
MixCvT-Base with the fully convolutional corner head outperforms that of the query-based head. In particular, MixCvT-Base with corner head surpass the state-of-the-art TransT trackers even without any post-processing and online templates. 
Besides, MixCvT-Base with the query head, a pure transformer-based tracking framework, obtains a much better performance than the query head of STARK-ST (66.0 vs 63.7).
These results demonstrate the generalization ability of our MAM backbone to different localization heads.

\noindent \textbf{MixViT's localization heads.}
To verify the effectiveness of our proposed pyramidal corner head on the MixViT tracker, we evaluate the performance of MixViT with plain corner head and pyramidal corner head as shown in Table~\ref{tab_mixvit_heads}. All trackers in Table~\ref{tab_mixvit_heads} are pre-trained with MAE encoders and deployed without the online template updating. When substituting plain corner head to pyramid corner head, the AUC score increases by 0.5. It indicates the effectiveness of the proposed pyramidal corner head, which can supplement the multi-scale information to the non-hierarchical MixViT backbone for better object localization.

\noindent \textbf{MixViT's positional embeddings.}
We evaluate the three types of positional embeddings of MixViT as referred in Section~\ref{sec_pos_emb}. The results are shown in Table~\ref{tab_mixvit_heads}. All trackers are pre-trained with MAE encoders and are deployed without online scheme. We can see that the MixViT models with the three different positional embeddings achieve comparable performance, which prove that it's no matter what initialization methods of positional embeddings are emplyed and whether to set it to be learnable.

\begin{table}[pt]
% \small
\caption{Study on pretraining methods. `Tracking Data' denotes for using the whole tracking datasets including GOT-10k, LaSOT, TrackingNet and COCO. `Scratch' means the model is trained from scratch. `MAE' with `Tracking data.' pre-training represents that we pretrain the MixViT backbone with MAE performed on tracking datasets including LaSOT, TrackingNet, GOT-10k and coco. {\bf All models employ plain corner head, asymmetric mechanism and are deployed without online scheme.}}
    \centering
    \fontsize{8.5pt}{4.5mm}\selectfont
    % \fontsize{8}{9}\selectfont  
    \setlength{\tabcolsep}{1.2mm}{
    % \resizebox{\linewidth}{!}{
    % \vspace{-1mm}
    % \resizebox{\columnwidth}{!}{
    % \small
    \begin{tabular}{c|cc|c}
        \hline
       Trackers & Pre-training Methods & Pre-training Data & AUC\\
       \hline
       MixCvT& From Scratch & - & 59.0 \\
       MixCvT& Supervised (CvT) & ImageNet-1k & 67.2 \\
        MixCvT& Supervised (CvT) & ImageNet-22k & \cellcolor{gray!20}\textbf{68.1} \\
        % MixCvT& ImageNet-22k Sup. & GOT-10k & 61.5 \\
       \hline
        \hline
        MixViT & From Scratch & - & 56.2 \\
        MixViT & MAE & Tracking Data. & 65.1 \\
        MixViT & TrackMAE & Tracking Data. & 66.0 \\
        MixViT & Supervised (DeiT3) & ImageNet-1k & 66.5 \\
        MixViT & Supervised (DeiT3) & ImageNet-21k & 67.7 \\
        MixViT & MAE Fine-tune & ImageNet-1k & 68.0 \\
        MixViT & MAE Pre-train & ImageNet-1k & \cellcolor{gray!20}\textbf{68.5} \\
        \hline
    \end{tabular}
    }
    % }
    \label{tab_pretraining}
% \vspace{-7mm}
\end{table}

\begin{figure}[pt]
\centering
\includegraphics[width=\linewidth]{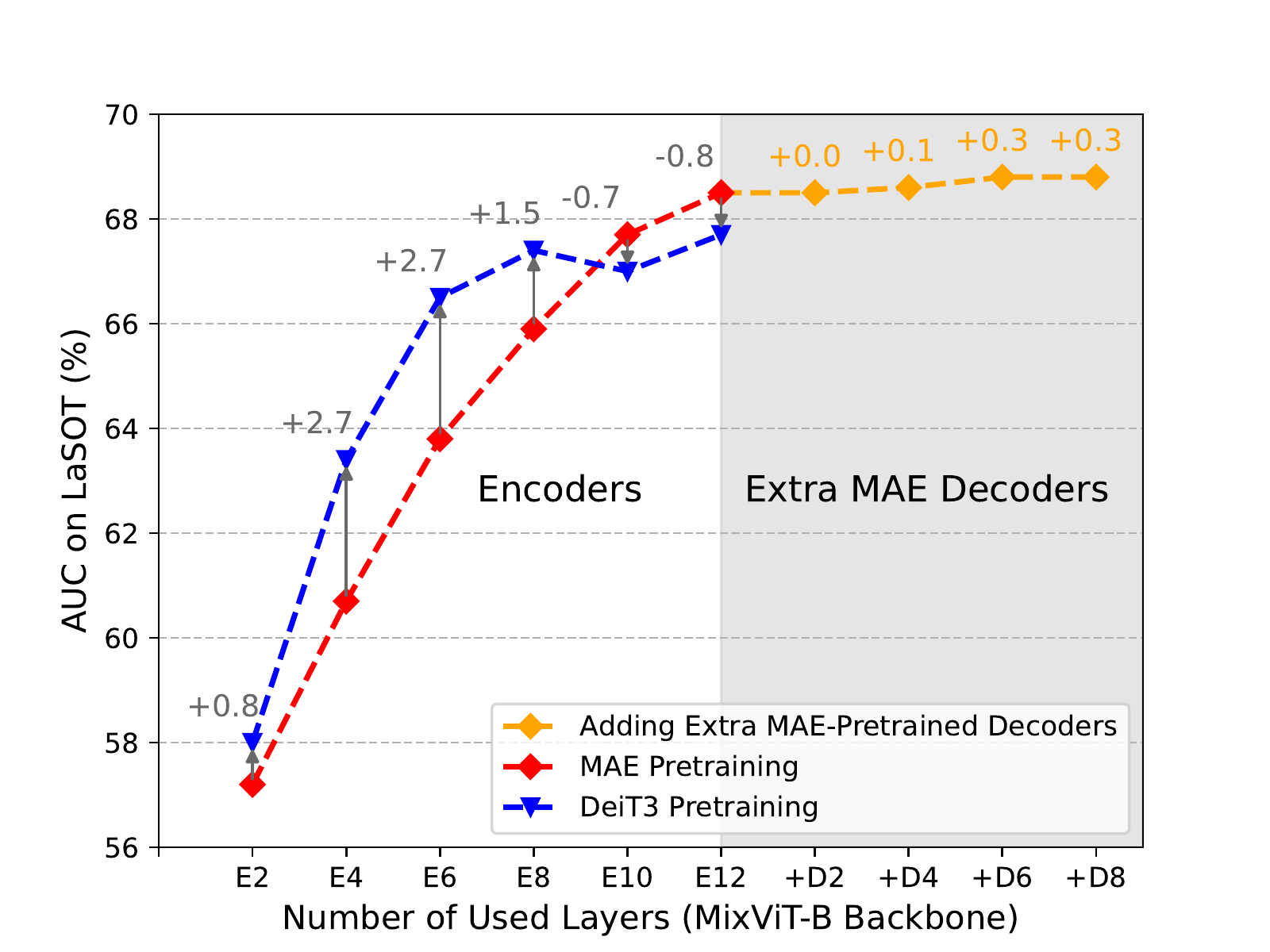}
\caption{The performance of AUC on LaSOT when we use MAE pretraining and DeiT3 pretraining on MixViT backbones with different number of layers. All models use plain corner head and are deployed without online scheme.}
\label{fig:ablation_pretraining}
\end{figure}

\subsubsection{Study on Pre-training Methods}
\label{section:pretraining}
\noindent \textbf{Study on different pre-training methods.}
Pre-training of MAM based backbone plays an important role in MixFormer tracking. We investigate some different pre-training methods of MixCvT and MixViT as shown in Table~\ref{tab_pretraining}.
Both MixCvT and MixViT with supervised pre-training or self-supervised pre-training outperform training from scratch by more than 9\%, which indicates the importance of pre-training for MixFormer trackers.
MixCvT and MixViT with ImageNet-21k pre-training obtain higher AUC than that with ImageNet-1k pre-training, which demonstrates the importance of semantic features trained with more supervision. 
MixViT with self-supervised pre-training of MAE outperforms that with the supervised pre-training of DeiT3.
The MixViT with tracking-data only MAE (trained with the tracking datasets instead of ImageNet) pre-training acquire comparable performance with that of supervised pre-training and largely surpass that trained from scratch. 
Meanwhile, our proposed TrackMAE surpasses the tracking-data only MAE by 0.9.
This demonstrates that it is meaningful to design a customized self-supervised pre-training pipeline for tracking framework, so as to narrow the gap between pre-training and fine-tuning.

\begin{table}[pt]
% \small
\caption{Study on online templates update mechanism and score prediction module. All models are pretrained on ImageNet-22k. Performance is evaluated on LaSOT.}
    \centering
    \fontsize{8.5pt}{4.5mm}\selectfont
    % \fontsize{8}{9}\selectfont  
    \setlength{\tabcolsep}{1.2mm}{
    % \resizebox{\linewidth}{!}{
    % \vspace{-1mm}
    % \resizebox{\columnwidth}{!}{
    % \small
    \begin{tabular}{c|cc|c}
        \hline
        Tracker & Online & Score of SPM &AUC\\
        \hline
        MixCvT& - & - & 68.1 \\
        MixCvT& \checkmark & - & 66.6 \\
        MixCvT& \checkmark & \checkmark & \cellcolor{gray!20}\textbf{69.2} \\
        \hline
    \end{tabular}
    }
    % }
    \label{tab_online}
% \vspace{-1mm}
\end{table}

\begin{table*}[ht]
    \centering
    \caption{State-of-the-art comparison on VOT2020~\cite{vot2020}. The best two results are shown in \textbf{\textcolor{red}{red}} and \textbf{\textcolor{blue}{blue}} fonts. Our trackers use Alpha-Refine~\cite{alpha-refine} to predict masks. MixFormer-1k is pretrained with ImageNet-1k.} 
    % \fontsize{7pt}{3.5mm}\selectfont
    % % \fontsize{7}{9}\selectfont  
    % \setlength{\tabcolsep}{0.6mm}{
    \resizebox{\linewidth}{!}{
    \begin{tabular}{c|ccccccccc|ccc|cc|cc}
    % \hline
    \toprule
         & SiamMask & D3S & SuperDiMP & AlphaRef & OceanPlus  & RPT & DualTFR & STARK & ToMP & \textbf{MixCvT-1k} & \textbf{MixCvT-22k} & \textbf{MixCvT-L} & \textbf{MixViT} & \textbf{MixViT-L} & \textbf{MixViT} & \textbf{MixViT-L}\cr
        & ~\cite{siammask} &~\cite{d3s} &~\cite{dimp} &~\cite{alpha-refine} &~\cite{oceanPlus} & ~\cite{rpt} & ~\cite{dualtfr}& ~\cite{stark} & ~\cite{tomp} & & & & & & (ConvMAE)& (ConvMAE) \cr
        % \hline
        \midrule
        EAO &  0.321 & 0.439 & 0.305 & 0.482 & 0.491 & 0.530 & 0.528 & 0.505 & 0.497 & 0.527 & 0.535 & \textbf{\textcolor{blue}{0.555}} & 0.518 & \textbf{\textcolor{red}{0.584}} & 0.528 & 0.567 \\
        Accuracy & 0.624 & 0.699 & 0.492 & 0.754 & 0.685 & 0.700 & 0.755 & 0.759 & 0.750 & 0.746 & \textbf{\textcolor{blue}{0.761}} & \textbf{\textcolor{red}{0.762}} & 0.728 & 0.755 & 0.734 & 0.747 \\
        Robustness & 0.648 & 0.769 & 0.745 & 0.777 & 0.842 & 0.869 & 0.836 & 0.817 & 0.798 &
        0.833 & 0.854 & 0.855 & 0.841 & \textbf{\textcolor{red}{0.890}} & 0.857 & \textbf{\textcolor{blue}{0.870}} \\
    % \hline
    \bottomrule
    \end{tabular}
    }
    % \vspace{1.5mm}
    \label{tab:vot2020}
% \vspace{-3mm}
\end{table*}

\begin{figure*}[pt]
\centering
\includegraphics[width=\linewidth]{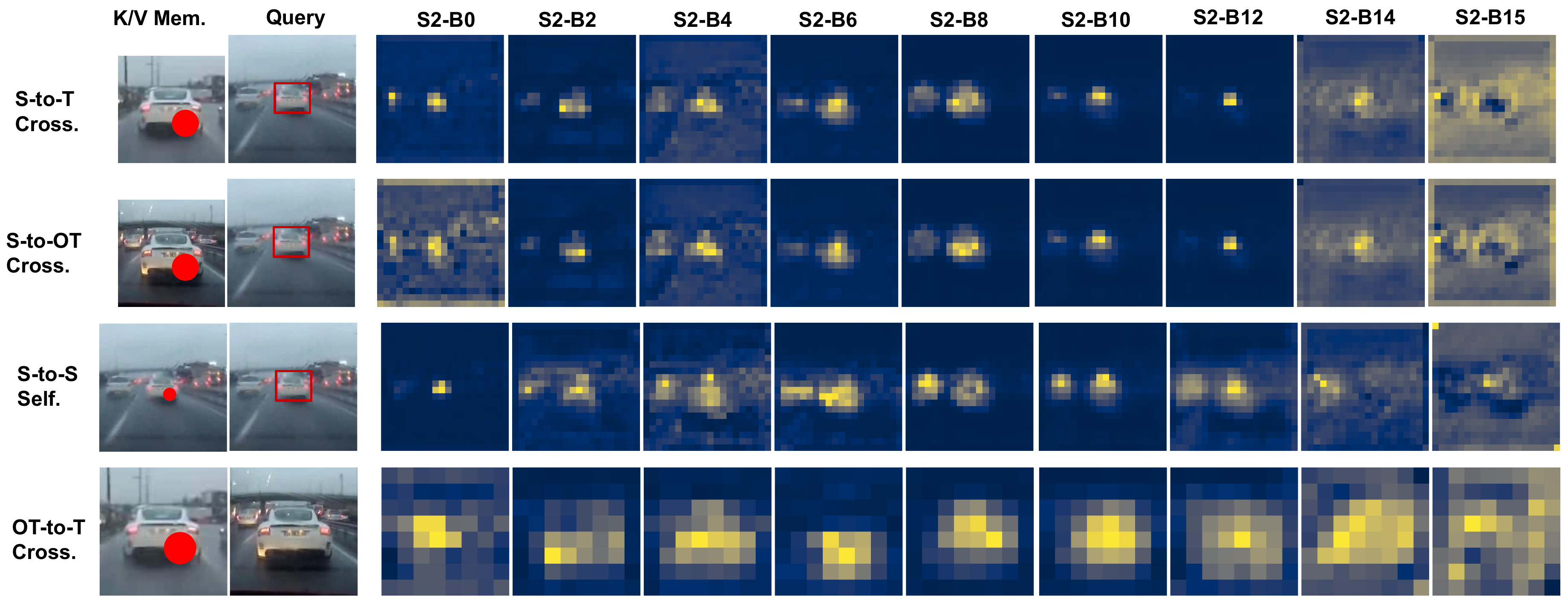}
% \vspace{-3mm}
\caption{Visualization results of different MixCvT's attention weights on \textit{car-2} of LaSOT. \textbf{S-to-t} is search-to-template cross attention, \textbf{S-to-OT} is search-to-online-template cross attention, \textbf{S-to-S} is self attention of search region and \textbf{OT-to-T} is online-template-to-template cross attention. \textbf{S$i$-B$j$} represents for Stage-$i$ and Block-$j$ of MixFormer. Best viewed with zooming in.}
% \vspace{-5mm}
\label{fig:vis_attn_car-2}
\end{figure*}

\begin{figure*}[pt]
\centering
\includegraphics[width=\linewidth]{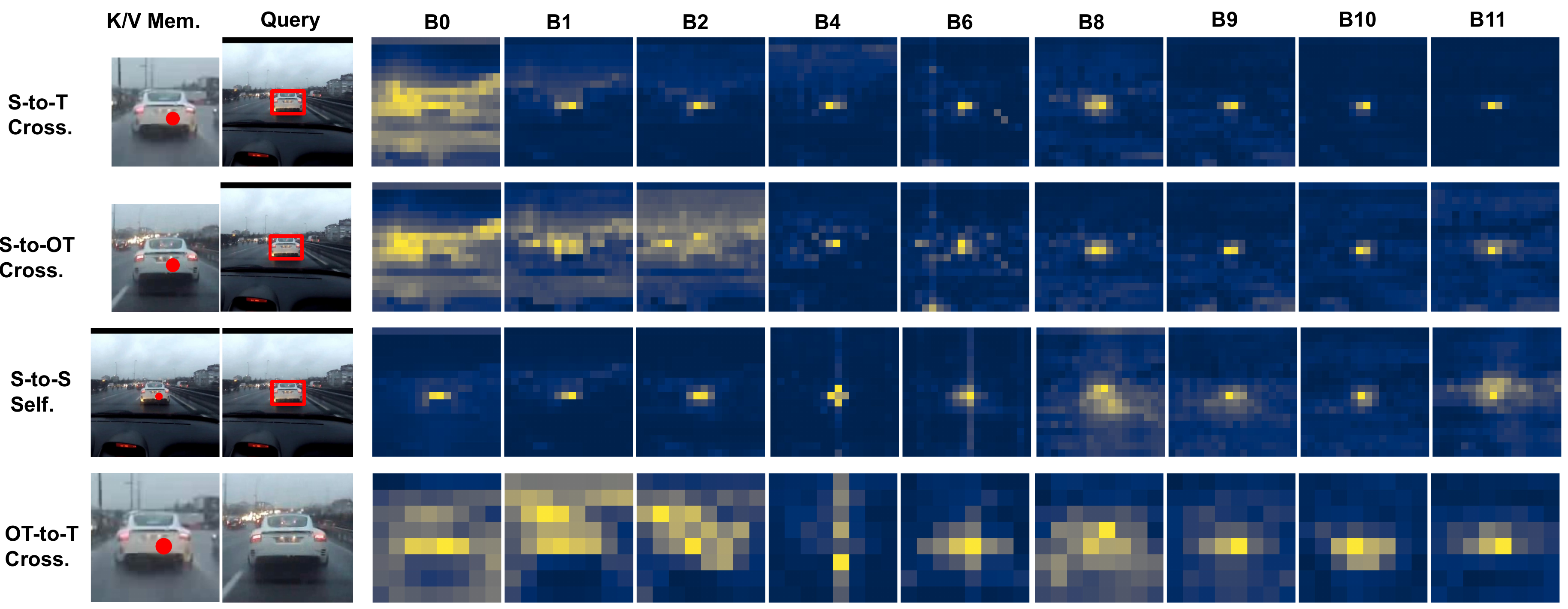}
% \vspace{-3mm}
\caption{Visualization results of different MixViT's attention weights on \textit{car-2} of LaSOT. \textbf{S-to-t} is search-to-template cross attention, \textbf{S-to-OT} is search-to-online-template cross attention, \textbf{S-to-S} is self attention of search region and \textbf{OT-to-T} is online-template-to-template cross attention. \textbf{S$i$-B$j$} represents for Stage-$i$ and Block-$j$ of MixFormer. Best viewed with zooming in.}
% \vspace{-5mm}
\label{fig:vis_attn_mixvit}
\end{figure*}

\noindent \textbf{Study on different number of encoder and decoder layers.}
To take a closer look at the difference between Deit3 pre-training and MAE pre-training, we investigate MixViT with different number of transformer layers using the two pre-training methods as shown in Fig.~\ref{fig:ablation_pretraining}.
 ``+D$N$" in X-axis of Fig.~\ref{fig:ablation_pretraining} represents for adding extra $N$ MAE decoders (with mixed attention).
For the supervised pre-training of DeiT3, MixViT of 8 layers reaches comparable performance with that of 12 layers. It indicates the supervised pre-training might learn high-level semantic features that have less relation with the object tracking.
However, for the self-supervised pre-training of MAE encoders, we see a different curve and the MixViT of 12 layers increases that of 8 layers by 2.6\%.
Furthermore, we add extra MAE decoders and see a small performance improvement.
We suppose that the last few layers provide very little help, since the representations are gradually adjusted to adapt to the pixel-level reconstruction task.
Finally, we surprisingly find that MixViT of 4 layers with DeiT3 pre-training still obtains a high AUC of 63.4\% with an extremely high running speed of 300 FPS.

% \noindent \textbf{Study on training and pre-training datasets of MixCvT.}
% To verify the generalization ablility of our MixCvT, we conduct an analysis on different pre-training and training datasets, as shown in Table~\ref{tab_pretraining}. MixCvT pretrained by ImageNet-1k still outperforms the SOTA trackers (e.g., TransT~\cite{tt}, KeepTrack~\cite{keeptrack}, STARK~\cite{stark}), even without post-processing and multi-layer feature aggregation. In addition, MixCvT trained with GOT-10k also achieves an impressive AUC of 62.1, which outperforms a majority of trackers trained with the whole tracking datasets.

\subsubsection{Study on online scheme and SPM}
We evaluate the effectiveness of the proposed Score Prediction Module and the online scheme.
As demonstrated in Table~\ref{tab_online}, MixCvT with online templates, sampled by a fixed update interval, performs worse than that with only the first template, and the online MixCvT with our score prediction module achieves the best AUC score. Specifically, the online MixCvT with our SPM improves the original MixCvT by 0.9\% of AUC on LaSOT. It suggests that selecting reliable templates with our score prediction module is of vital importance, since low-quality online samples can bring noises to the templates. 

% \vspace{-4mm}

% \vspace{-4mm}
\subsection{Visualization Results}
% \noindent \textbf{Visualization of attention maps.}
\label{vis_attn}
To explore how the mixed attention works in our MixFormer trackers, we visualize the attention maps of MixCvT in Fig.~\ref{fig:vis_attn_car-2} and the attention maps of MixViT in Fig.~\ref{fig:vis_attn_mixvit}. 
From the four types of attention maps, we derive that: (\romannumeral1) distractors in background get suppressed layer by layer, demonstrating the effectiveness of mixed attention on enhancing the tracker's discriminative ability; (\romannumeral2) online templates are supplements to the static template, which may be more adaptive to appearance variation and help to discriminate the target from the background; (\romannumeral3) the foreground of multiple templates can be augmented by mutual cross attention (from the fourth line of `OT-to-T Cross' visualization), (\romannumeral4) a certain position tends to interact with the surrounding local patch (from the `S-to-S Self.' visualization). (\romannumeral5) compared with the visualization results of MixCvT, MixViT yields a cleaner attention maps, especially in the last blocks. We think this is possibly due to the different pre-training methods. 

\begin{table*}[pt]
\caption{State-of-the-art comparison on TrackingNet~\cite{trackingnet}, LaSOT~\cite{lasot}, GOT-10k~\cite{got10k} and UAV123~\cite{uav123}. The best two results are shown in \textbf{\textcolor{red}{red}} and \textbf{\textcolor{blue}{blue}} fonts. The \underline{underline} results of GOT-10k are not considered in the comparison, since the models are trained with datasets other than GOT-10k. MixCvT-1k is the model pre-trained with ImageNet-1k. Other MixCvT models are pre-trained on ImageNet-22k and all MixViT models are pre-trained with corresponding MAE-pretraining encoders. `*' denotes for trackers trained only with GOT-10k train split.}
    \centering
    % \small
    % \fontsize{7pt}{3.5mm}\selectfont
    % \fontsize{8}{9}\selectfont  
    % \setlength{\tabcolsep}{2.0mm}{
    \resizebox{\linewidth}{!}{
    \begin{tabular}{c|ccc|ccc|ccc|cc}
    \toprule
    % \hline
    \multirow{2}{*}{Method} &
    \multicolumn{3}{c|}{LaSOT} &
    \multicolumn{3}{c|}{TrackingNet} &
    \multicolumn{3}{c|}{GOT-10k} &
    \multicolumn{2}{c}{UAV123} \\
    \cline{2-12}
     & AUC(\%) & $P_{Norm}$(\%) & P(\%) & AUC(\%) & $P_{Norm}(\%)$ & P(\%) & AO(\%) & $SR_{0.5}$(\%) & $SR_{0.75}$(\%) & AUC(\%) & P(\%)\\
    \midrule
    \textbf{MixViT-L (ConvMAE)} & \textbf{\textcolor{red}{73.3}} &  
 \textbf{\textcolor{red}{82.8}} & \textbf{\textcolor{red}{80.3}} & \textbf{\textcolor{red}{86.1}} & \textbf{\textcolor{red}{90.3}} & \textbf{\textcolor{red}{86.0}} &
    \underline{75.4} & \underline{84.0} & \underline{75.4} & 70.0
     &  \textbf{\textcolor{red}{92.5}} \\
    \textbf{MixViT (ConvMAE)} & 70.4 & 80.4 & 76.7 & 84.5 & 89.1 & 83.7 &
    \underline{74.3} & \underline{84.1} & \underline{73.0} & 69.0
     & 91.2 \\
    \midrule
    % \hline
   \textbf{MixViT-L} & \textbf{\textcolor{blue}{72.4}} &  
 \textbf{\textcolor{blue}{82.2}} & \textbf{\textcolor{blue}{80.1}} & \textbf{\textcolor{blue}{85.4}} & \textbf{\textcolor{blue}{90.2}} & \textbf{\textcolor{blue}{85.7}} &
    \underline{78.0} & \underline{87.3} & \underline{78.7} & 68.7
     & 90.8 \\
    \textbf{MixViT} & 69.6 & 79.9 & 75.9 & 83.5 & 88.3 & 82.0 &
    \underline{72.7} & \underline{82.3} & \underline{70.8} & 68.1
     & 90.3 \\
    \textbf{MixViT-L*} & - & - & - & - & - & - & \textbf{\textcolor{red}{75.7}} & \textbf{\textcolor{red}{85.3}} & \textbf{\textcolor{red}{75.1}} & - & - \\ 
    \textbf{MixViT*} & - & - & - & - & - & - & 72.5 & 82.4 & 69.9 & - & - \\ 
    \midrule
    \textbf{MixCvT-L} & 70.1 & 79.9 & 76.3 & 83.9 & 88.9 & 83.1 &
    \underline{75.6} & \underline{85.7} & \underline{72.8} &
    69.5 & 91.0 \\
    \textbf{MixCvT} & 69.2 & 78.7 & 74.7 & 83.1 & 88.1 & 81.6
    & \underline{72.6} & \underline{82.2} & \underline{68.8} & 70.4 & \textbf{\textcolor{blue}{91.8}} \\
    \textbf{MixCvT-1k} & 67.9 & 77.3 & 73.9 & 82.6 & 87.7 & 81.2 & \underline{73.2} & \underline{83.2} & \underline{70.2} & 68.7 & 89.5 \\ 
     \textbf{MixCvT*} & - & - & - & - & - & - & 70.7 & 80.0 & 67.8 & - & - \\ 
    \textbf{MixCvT-1k*} & - & - & - & - & - & - & 71.2 &  79.9 &  65.8 & - & - \\ 
    \midrule
    SimTrack-L~\cite{simtrack} & 70.5 & 79.7 & - & 83.4 & 87.4 & - & 69.8 & 78.8 & 66.0 & \textbf{\textcolor{red}{71.2}} & 91.6 \\
    SimTrack-B~\cite{simtrack} & 69.3 & 78.5 & - & 82.3 & 86.5 & - & 68.6 & 78.9 & 62.4 & 69.8 & 89.6 \\
    OSTrack-384~\cite{ostrack} & 71.1 & 81.1 & 77.6 & 83.9 & 88.5 & 83.2 & \textbf{\textcolor{blue}{73.7}} & \textbf{\textcolor{blue}{83.2}} & \textbf{\textcolor{blue}{70.8}} & \textbf{\textcolor{blue}{70.7}} & - \\
    OSTrack-256~\cite{ostrack} & 69.1 & 78.7 & 75.2 & 83.1 & 87.8 & 82.0 & 71.0 & 80.4 & 68.2 & 68.3 & - \\
    AiATrack~\cite{aiatrack} & 69.0 & 79.4 & 73.8 & 82.7 & 87.8 & 80.4 & 69.6 & 80.0 & 63.2 & 70.6 & - \\
    SwinTrack-B-384~\cite{swintrack} & 71.3 & - & 76.5 & 84.0 & - & 82.8 & 72.4 & 80.5 & 67.8 & - & - \\
    ToMP-101~\cite{tomp} & 68.5 & 79.2 & 73.5 & 81.5 & 86.4 & 78.9 & - & - & - & 66.9 & - \\
    CSWinTT~\cite{cswintt} & 66.2 & 75.2 & 70.9 & 81.9 & 86.7 & 79.5 & 69.4 & 78.9 & 65.4 & 70.5 & 90.3 \\
    SBT-large~\cite{sbt} & 66.7 & - & 71.1 & - & - & - & 70.4 & 80.8 & 64.7 & - & - \\
    UTT~\cite{utt} & 64.6 & - & 67.2 & 79.7 & - & 77.0 & 67.2 & 76.3 & 60.5 & - & - \\
    STARK~\cite{stark} & 67.1 & 77.0 & - & 82.0 & 86.9 & - & 68.8 & 78.1 & 64.1 & - & - \\
    KeepTrack~\cite{keeptrack} & 67.1 & 77.2 & 70.2 & - & - & - & - & - & - & 69.7 & - \\
    DTT~\cite{dtt} & 60.1 & - & - & 79.6 & 85.0 & 78.9 & 63.4 & 74.9 & 51.4 & - & - \\
    SAOT~\cite{saot} & 61.6 & 70.8 & - & - & - & - & 64.0 & 75.9 & - & - & - \\
    AutoMatch~\cite{automatch} & 58.2 & - & 59.9 & 76.0 & - & 72.6 & 65.2 & 76.6 & 54.3 & - & - \\
    TREG~\cite{treg} & 64.0 & 74.1 & - & 78.5 & 83.8 & 75.0 & 66.8 & 77.8 & 57.2 & 66.9 & 88.4 \\
    DualTFR~\cite{dualtfr} & 63.5 & 72.0 & 66.5 & 80.1 & 84.9 & - & - & - & - & 68.2 & - \\
    TransT~\cite{transt} & 64.9 & 73.8 & 69.0 & 81.4 & 86.7 & 80.3 & 67.1 & 76.8 & 60.9 & 69.1 & - \\
    TrDiMP~\cite{tmt} & 63.9 & - & 61.4 & 78.4 & 83.3 & 73.1 & 67.1 & 77.7 & 58.3 & 67.5 & - \\
    STMTracker~\cite{stmtrack} & 60.6 & 69.3 & 63.3 & 80.3 & 85.1 & 76.7 & 64.2 & 73.7 & 57.5 & 64.7 & - \\
    SiamR-CNN~\cite{siamrcnn} & 64.8 & 72.2 & - & 81.2 & 85.4 & 80.0 & 64.9 & 72.8 & 59.7 & 64.9 & 83.4 \\
    PrDiMP~\cite{prdimp} & 59.8 & 68.8 & 60.8 & 75.8 & 81.6 & 70.4 & 63.4 & 73.8 & 54.3 & 68.0 & - \\
    OCEAN~\cite{ocean} & 56.0 & 65.1 & 56.6 & - & - & - & 61.1& 72.1 & 47.3 & - & - \\
    KYS~\cite{kys} & 55.4 & 63.3 & - & 74.0 & 80.0 & 68.8 &  63.6 & 75.1 & 51.5 & - & - \\
    FCOT~\cite{fcot} & 57.2 & 67.8 & - & 75.4 & 82.9 & 72.6 & 63.4 & 76.6 & 52.1 & 65.6 & 87.3 \\
    SiamAttn~\cite{siamattn} & 56.0 & 64.8 & - & 75.2 & 81.7 & - & - & - & - & 65.0 & 84.5 \\
    SiamGAT~\cite{siamgat} & 53.9 & 63.3 & 53.0 & - & - & - & 62.7 & 74.3 & 48.8 & 64.6 & 84.3 \\
    CGACD~\cite{CGACD} & 51.8 & 62.6 & - & 71.1 & 80.0 & 69.3 & - & - & - & 63.3 & 83.3 \\
    SiamBAN~\cite{siamban} & 53.1 & - &  54.1 & 71.6  & 79.4  & 68.5  & - & - & - & 64.4  &  84.6 \\
    SiamCAR~\cite{siamcar} & 50.7 & 60.0 &  51.0 & -  & - & - & 56.9 & 67.0 & 41.5 & 61.4 & 76.0 \\
    SiamFC++~\cite{siamfc++} & 54.4& 62.3 & 54.7 & 75.4 & 80.0 & 70.5 & 59.5& 69.5 & 47.9 & - & - \\
    MAML~\cite{maml} & 52.3 & - & - & 75.7 & 82.2 & 72.5 & - & - & - & - & - \\
    D3S~\cite{d3s} & - & - & - & 72.8 & 76.8 & 66.4 & 59.7 & 67.6 & 46.2 & - & - \\
    DiMP~\cite{dimp} & 56.9 & 65.0 & 56.7 & 74.0 & 80.1 & 68.7 & 61.1 & 71.7 & 49.2 & 65.4 & - \\
    ATOM~\cite{atom} & 51.5 & 57.6 & 50.5 & 70.3 & 77.1 & 64.8 & 55.6 & 63.4 & 40.2 & 64.3 & - \\
    SiamRPN++~\cite{siamrpnPlus} & 49.6 & 56.9 & 49.1 & 73.3 & 80.0 & 69.4 & 51.7 & 61.6 & 32.5 & 61.0 & 80.3 \\
    SiamMask~\cite{siammask} & - & - & - & 72.5 & 77.8 & 66.4 & 51.4 & 36.6 & 58.7 & - & - \\
    ECO~\cite{eco}& 32.4 & 33.8 & 30.1 & 55.4 & 61.8 & 49.2 & 31.6 & 30.9 & 11.1 & 53.2 & - \\
    MDNet~\cite{mdnet} & 39.7 & 46.0 & 37.3 & 60.6 & 70.5 & 56.5 & 29.9 & 30.3 & 9.9 & 52.8 & - \\
    SiamFC~\cite{siamfc} & 33.6 & 42.0 & 33.9 & 57.1 & 66.3 & 53.3 & 34.8 & 35.3 & 9.8 & 48.5 & 69.3 \\
    % \hline
      \bottomrule
    \end{tabular}}
    % }
    % \vspace{-5mm}
    \label{tab:resultsl}
\end{table*}

\subsection{Comparison with the state-of-the-art trackers}
We verify the performance of our proposed MixCvT-1k, MixCvT-22k, MixCvT-L, MixViT and MixViT-L on six benchmarks, including VOT2020~\cite{vot2020}, LaSOT~\cite{lasot}, TrackingNet~\cite{trackingnet}, GOT10k~\cite{got10k}, UAV123~\cite{uav123} and OTB100~\cite{otb}.
% \vspace{-4mm}

\begin{figure*}[pt]
\centering
\includegraphics[width=\linewidth]{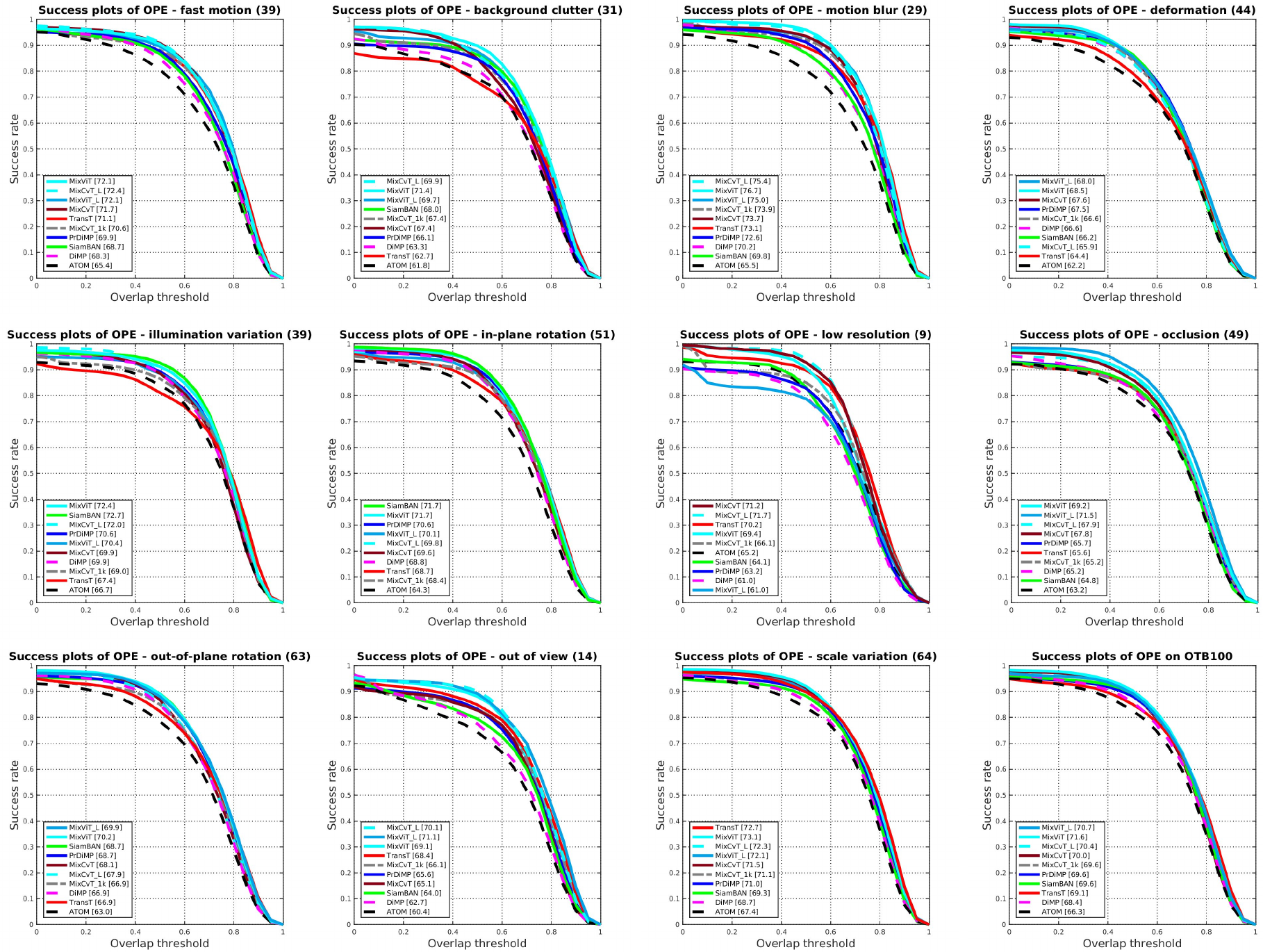}
% \vspace{-8mm}
\caption{Comparisons on OTB100~\cite{otb} with different challenging aspects: fast motion, background clutter, motion blur, deformation, illumination variation, in-plane rotation, low resolution, occlusion, out-of-plane rotation, out of view and scale variation.}
% \vspace{-5mm}
\label{fig:detailed_otb}
\end{figure*}

\begin{figure}[pt]
\centering
\includegraphics[width=\linewidth]{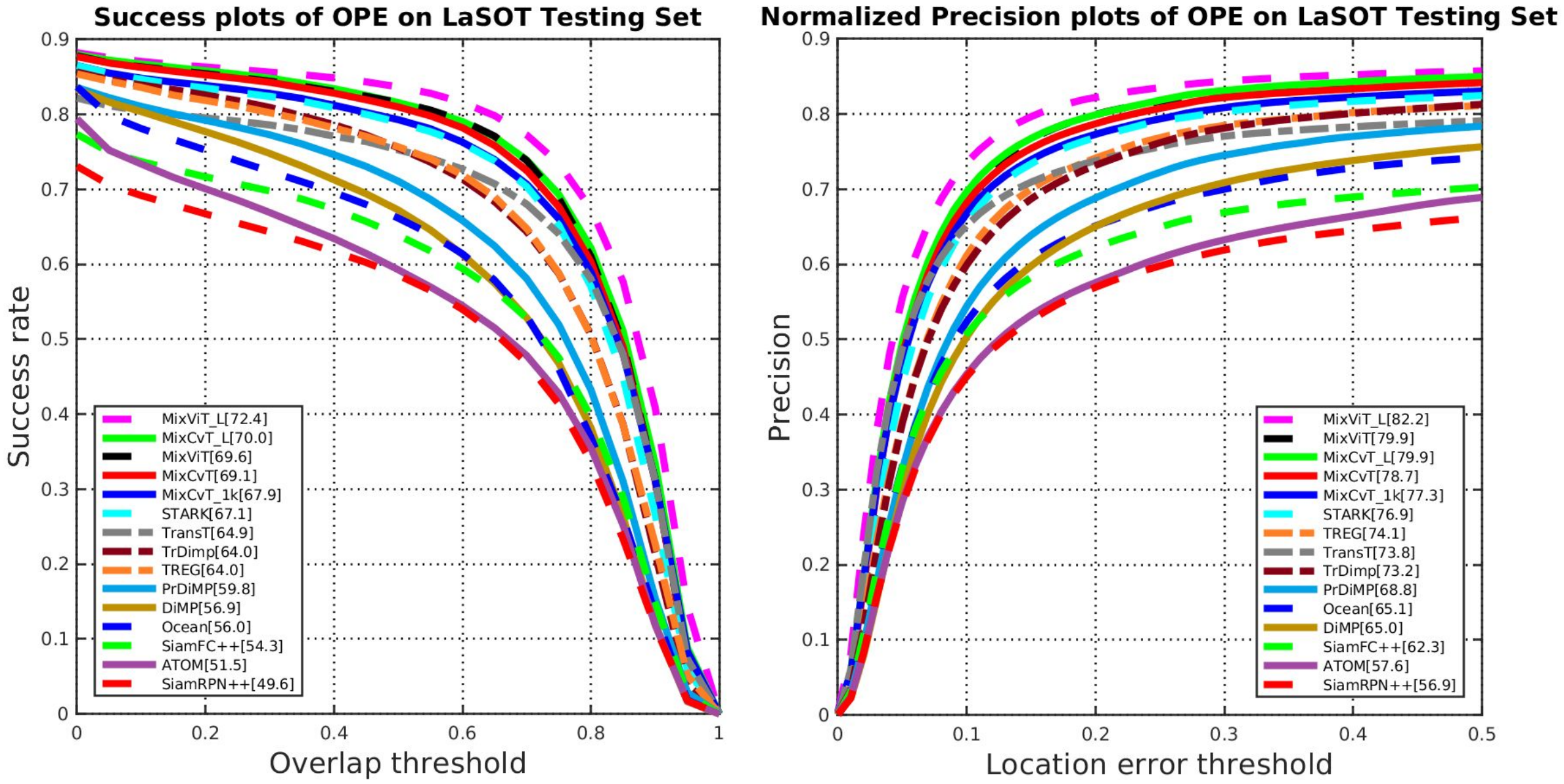}
% \vspace{-8mm}
\caption{State-of-the-art comparison on the LaSOT dataset. Best viewed with zooming in.}
% \vspace{-5mm}
\label{fig:lasot}
\end{figure}

\begin{figure}[pt]
\centering
\includegraphics[width=\linewidth]{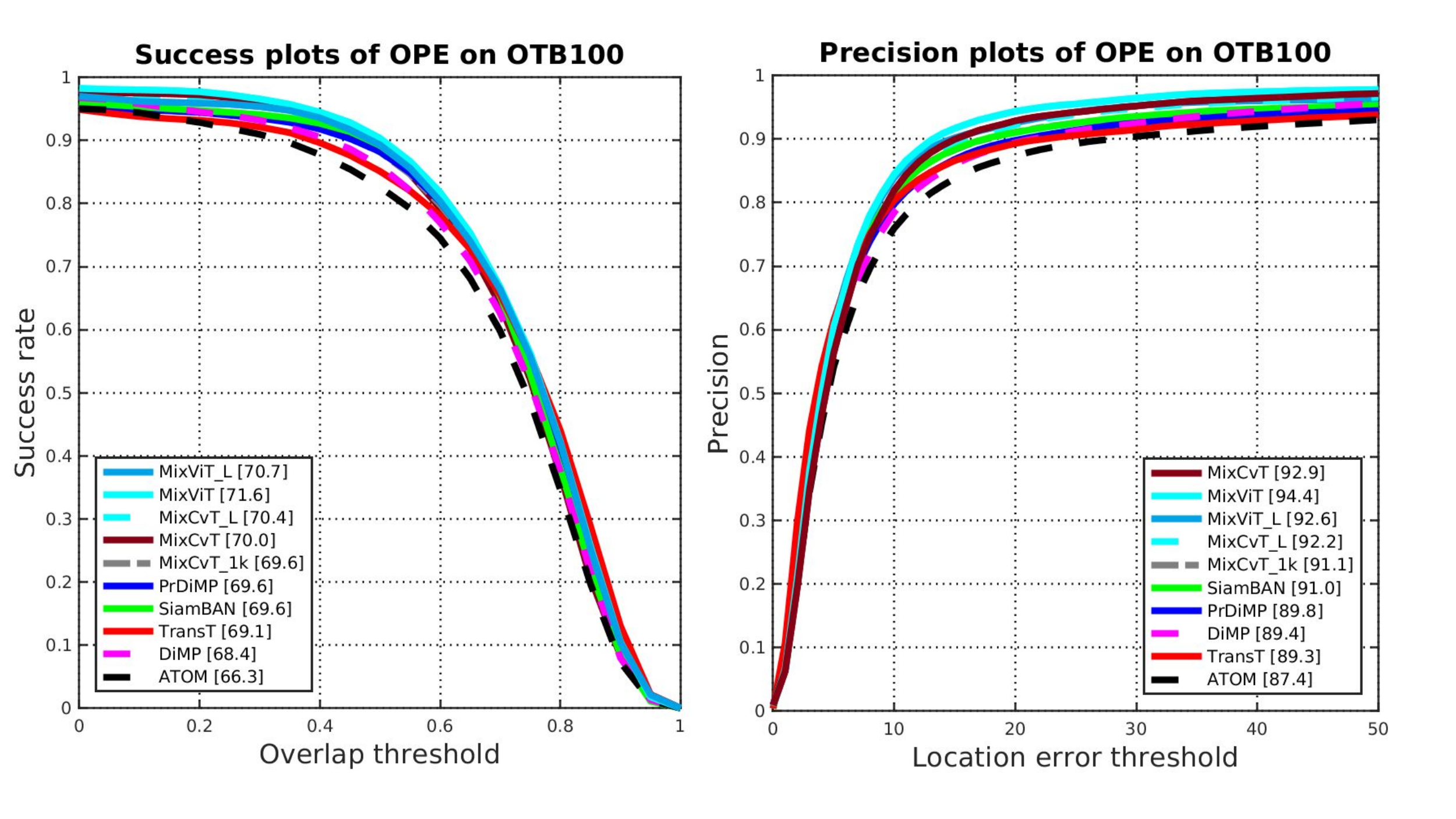}
\caption{State-of-the-art comparison on the OTB100 dataset. Best viewed with zooming in.}
\label{fig:otb}
\end{figure}

\noindent \textbf{VOT2020.}
VOT2020~\cite{vot2020} consists of 60 videos with several challenges including fast motion, occlusion, etc.
Our trackers are tested on the dataset in comparison with state-of-the-art trackers.
As shown in Table~\ref{tab:vot2020}, MixViT-L achieves the top-ranked performance on EAO criteria of 0.584, which outperforms the transformer tracker STARK with a large margin of 7.9\% of EAO. MixCvT-22k and MixViT-L outperform other trackers including RPT (VOT2020 short-term challenge winner). Besides, MixViT-L with ConvMAE pre-training obtains lower EAO than MixViT-L. We analyze that this is mainly because MixViT-L with ConvMAE pre-training removes the last three layers of original ConvMAE which may lack some high-level representation and thus yielding a negative impact on Robustness.

% \vspace{-4mm}
\noindent \textbf{LaSOT.}
LaSOT~\cite{lasot} has 280 videos in its test set. We evaluate our MixFormer on the test set to validate its long-term capability. 
The Table~\ref{tab:resultsl} shows that our MixFormer models surpass all other trackers with a large margin. Specifically, MixViT-L pretrained with ConvMAE achieves the top-ranked performance on AUC of 73.3\%, surpassing ToMP~\cite{tomp} by 4.8\% and SimTrack-L~\cite{simtrack} (based on ViT-Large) by 2.8\%. To give a further analysis, we provide Success plot and Precision plot for LaSOT in Fig.~\ref{fig:lasot}. It proves that improvement is due to both higher accuracy and robustness. 

% \vspace{-4mm}
\noindent \textbf{TrackingNet.}
TrackingNet~\cite{trackingnet} provides over 30K videos with more than 14 million dense bounding box annotations. The videos are sampled from YouTube, covering target categories and scenes in real life. We validate our MixFormer models on its test set. From Table~\ref{tab:resultsl}, we find that our MixCvT-L and MixViT-L pretrained with ConvMAE set a new state-of-the-art performance on the large scale benchmark, with AUC of 83.9\% and 86.3\% respectively.
The results indicate that the proposed framework is effective for tracking.
% \vspace{-4mm}

\noindent \textbf{GOT10k.}
GOT10k~\cite{got10k} is a large-scale dataset with over 10000 video segments and has 180 segments for the test set. Apart from generic classes of moving objects and motion patterns, the object classes in the train and test set are zero-overlapped. As shown in Table~\ref{tab:resultsl}, our MixViT-L obtains state-of-the-art performance on the test split, outperforming SBT-large by 5.3\%. MixViT-L pretrained with ConvMAE-L obtains lower AO than MixViT-L pretrained with MAE-L due to the reason similar with that of VOT2020.
% \vspace{-4mm}

\noindent \textbf{UAV123.}
UAV123~\cite{uav123} is a large dataset containing 123 Sequences with average sequence length of 915 frames, which is captured from low-altitude UAVs. Table~\ref{tab:resultsl} shows our results on UAV123 dataset. 
MixViT-L and MixCvT achieve competitive performance with the ToMP-101~\cite{tomp} and CSWinTT~\cite{cswintt}.
% \vspace{-4mm}

\begin{table}[pt]
% \small
\caption{The Top eight trackers for VOT-STb2022 public challenge~\cite{vot22}. Expected average overlap (EAO), accuracy and robustness are shown. }
    \centering
    \fontsize{8.5pt}{5mm}\selectfont
    % \fontsize{8}{9}\selectfont  
    \setlength{\tabcolsep}{1.2mm}{
    % \resizebox{\linewidth}{!}{
    \vspace{-2mm}
    % \resizebox{\columnwidth}{!}{
    % \small
    \begin{tabular}{c|ccc}
        \hline
        Tracker & EAO & Accuracy & Robustness\\
        \hline
        MixFormerL & \textbf{0.602} & \textbf{0.831} & 0.859 \\
        DAMT & 0.602 & 0.776 & \textbf{0.887} \\
        OSTrackSTB & 0.591 & 0.790 & 0.869\\
        APMT\_MR & 0.591 & 0.787 & 0.877 \\
        MixFormer & 0.587 & 0.797 & 0.874 \\
        APMT\_RT & 0.581 & 0.787 & 0.877 \\
        ADOTstb & 0.569 & 0.775 & 0.862 \\
        SRATransT & 0.560 & 0.764 & 0.864 \\
        \hline
    \end{tabular}
    }
    % }
    \label{tab_vot}
\vspace{-1mm}
\end{table}

\begin{table}[pt]
% \small
\caption{The Top eight trackers for VOT-RGBD2022 challenge~\cite{vot22}. Expected average overlap (EAO), accuracy and robustness are shown. }
    \centering
    \fontsize{8.5pt}{5mm}\selectfont
    % \fontsize{8}{9}\selectfont  
    \setlength{\tabcolsep}{1.2mm}{
    % \resizebox{\linewidth}{!}{
    \vspace{-2mm}
    % \resizebox{\columnwidth}{!}{
    % \small
    \begin{tabular}{c|ccc}
        \hline
        Tracker & EAO & Accuracy & Robustness\\
        \hline
        MixForRGBD & \textbf{0.779} & \textbf{0.816} & \textbf{0.946} \\
        SAMF & 0.762 & 0.807 & 0.936 \\
        OSTrack & 0.729 & 0.808 & 0.894\\
        ProMix & 0.722 & 0.798 & 0.900 \\
        SBT\_RGBD & 0.708 & 0.809 & 0.864 \\
        DMTracker & 0.658 & 0.758 & 0.851 \\
        DeT\_DiMP50\_Max & 0.657 & 0.760 & 0.845 \\
        SPT & 0.651 & 0.798 & 0.851 \\
        \hline
    \end{tabular}
    }
    % }
    \label{tab_vot_rgbd}
\vspace{-1mm}
\end{table}

\noindent \textbf{OTB-100.}
OTB100~\cite{otb} is a commonly used benchmark, which evaluates performance on Precision and AUC scores. Figure.~\ref{fig:otb} presents results of our trackers on both the two metrics. MixCvT-L reaches competitive performance w.r.t. state-of-the-art trackers, surpassing the transformer tracker TransT by 1.3\% on AUC score. MixViT with MAE pre-training obains the highest AUC score of 71.6\% and Precision score of 94.4\%. 
In Figure.~\ref{fig:detailed_otb}, we compare MixFormer models with prevailing trackers on some different challenging aspects. The results prove that a series of MixCvT and MixViT models consistently outperform other trackers such as TransT~\cite{transt}, SiamBAN~\cite{siamban} and PrDiMP~\cite{prdimp} on most of the challenging aspects. 
Especially, MixViT significantly improves TransT~\cite{transt} by 8.7\% and 3.6\% on the aspects of background clutter and motion blur respectively, which indicates the strong discriminative ability of the proposed mixed attention.

\begin{table}[pt]
% \small
\caption{The Top six trackers for VOT-D2022 challenge~\cite{vot22}. Expected average overlap (EAO), accuracy and robustness are shown. }
    \centering
    \fontsize{8.5pt}{5mm}\selectfont
    % \fontsize{8}{9}\selectfont  
    \setlength{\tabcolsep}{1.2mm}{
    % \resizebox{\linewidth}{!}{
    \vspace{-2mm}
    % \resizebox{\columnwidth}{!}{
    % \small
    \begin{tabular}{c|ccc}
        \hline
        Tracker & EAO & Accuracy & Robustness\\
        \hline
        MixFormerD & \textbf{0.600} & \textbf{0.758} & \textbf{0.806} \\
        RSDiMP & 0.573 & 0.734 & 0.759 \\
        OSTrack\_D & 0.568 & 0.735 & 0.774\\
        DOT & 0.469 & 0.672 & 0.673 \\
        SBT\_Depth & 0.462 & 0.756 & 0.571 \\
        UpDoT & 0.439 & 0.652 & 0.627 \\
        % CoDeT & 0.372 & 0.597 & 0.594 \\
        % DiMP & 0.336 & 0.623 & 0.496 \\
        \hline
    \end{tabular}
    }
    % }
    \label{tab_vot_d}
\vspace{-1mm}
\end{table}

\noindent \textbf{VOT2022 Challenges Results.}
We also test the performance of our MixFormer trackers in the VOT2022 challenge. As shown in Table~\ref{tab_vot}, ours MixFormerL (based on the MixViT-L tracker) ranks 1/41 on the VOT2022-STb public challenge.
Besides, we notice that the VOT2022-RGBD and VOT2022-D winners of MixForRGBD and MixForD, implemented by Lai Simiao, are both constructed upon our MixFormer. This demonstrates the effectiveness and generalization of the proposed MixFormer framework.

\section{Conclusions}
We have presented MixFormer, an compact end-to-end tracking framework with iterative mixed attentions, aiming to unify the feature extraction and target integration. Mixed attention module performs both feature extraction and mutual interaction for target template and search area. 
We instanitate two types of MixFormer trackers: a hierarchical tracker of MixCvT and a non-hierarchical tracker of MixViT. We have performed extensive study on the design and pre-training techniques of MixFormer trackers. We also extend the MAE pre-training to the tracking framework and propose the TrackMAE to achieve the competitive performance to the large-scale ImageNet pre-training.
Extensive evaluations on six common tracking benchmarks demonstrates that our MixFormer tracker obtains a notable improvement over other prevailing trackers.
In the future, we consider extending MixFormer to multiple object tracking or long-term object tracking.

\ifCLASSOPTIONcompsoc
  \section*{Acknowledgments}
\else
  \section*{Acknowledgment}
\fi

This work is supported by National Natural Science Foundation of China  (No.62076119, No.61921006),  Program for Innovative Talents and Entrepreneur in Jiangsu Province, and Collaborative Innovation Center of Novel Software Technology and Industrialization.

\ifCLASSOPTIONcaptionsoff
  \newpage
\fi

{\small
\bibliographystyle{ieee_fullname}
\bibliography{egbib}
}

\end{document}